\documentclass[sigconf]{acmart}
\AtBeginDocument{%
  }

\setcopyright{acmlicensed}
\copyrightyear{2018}
\acmYear{2018}
\acmDOI{XXXXXXX.XXXXXXX}
\acmConference[Conference acronym 'XX]{Make sure to enter the correct
  conference title from your rights confirmation email}{June 03--05,
  2018}{Woodstock, NY}
\acmISBN{978-1-4503-XXXX-X/2018/06}

\usepackage{graphicx}
\usepackage{subcaption} 
\usepackage{enumitem}
\usepackage{multirow}
\usepackage{booktabs}
\usepackage{balance}
\usepackage{stfloats}
\usepackage{xcolor}

\begin{document}

\title{Defining and Benchmarking a Data-Centric Design Space \\for Brain Graph Construction}
\renewcommand{\shorttitle}{Defining and Benchmarking a Data-Centric Design Space \\for Brain Graph Construction}

\author{Qinwen Ge}
\affiliation{%
 \institution{Vanderbilt University}
 \city{Nashville}
 \state{TN}
 \country{USA}
 }
\email{qinwen.ge@vanderbilt.edu}

 \author{Roza G. Bayrak}
 \affiliation{%
 \institution{Vanderbilt University}
  \city{Nashville}
 \state{TN}
 \country{USA}
 }
\email{roza.g.bayrak@vanderbilt.edu}

\author{Anwar Said}
\affiliation{%
 \institution{Vanderbilt University}
  \city{Nashville}
 \state{TN}
 \country{USA}
 }
\email{anwar.said@vanderbilt.edu}

   \author{Catie Chang}
   \affiliation{%
 \institution{Vanderbilt University}
  \city{Nashville}
 \state{TN}
 \country{USA}
 }
\email{catie.chang@vanderbilt.edu}

\author{Xenofon Koutsoukos}
\affiliation{%
 \institution{Vanderbilt University}
  \city{Nashville}
 \state{TN}
 \country{USA}
 }
\email{xenofon.koutsoukos@vanderbilt.edu}

 \author{Tyler Derr}
 \affiliation{%
 \institution{Vanderbilt University}
  \city{Nashville}
 \state{TN}
 \country{USA}
 }
\email{tyler.derr@vanderbilt.edu}

\renewcommand{\shortauthors}{Trovato et al.}

\begin{abstract}
The construction of brain graphs from functional Magnetic Resonance Imaging (fMRI) data plays a crucial role in enabling graph machine learning for neuroimaging. However, current practices often rely on rigid pipelines that overlook critical data-centric choices in how brain graphs are constructed. In this work, we adopt a Data-Centric AI perspective and systematically define and benchmark a data-centric design space for brain graph construction, constrasting with primarily model-centric prior work. We organize this design space into three stages: temporal signal processing, topology extraction, and graph featurization. Our contributions lie less in novel components and more in evaluating how combinations of existing and modified techniques influence downstream performance. Specifically, we study high-amplitude BOLD signal filtering, sparsification and unification strategies for connectivity, alternative correlation metrics, and multi-view node and edge features, such as incorporating lagged dynamics. Experiments on the HCP1200 and ABIDE datasets show that thoughtful data-centric configurations consistently improve classification accuracy over standard pipelines. These findings highlight the critical role of upstream data decisions and underscore the importance of systematically exploring the data-centric design space for graph-based neuroimaging. Our code is available at \href{https://github.com/GeQinwen/DataCentricBrainGraphs}{https://github.com/GeQinwen/DataCentricBrainGraphs}.

\end{abstract}

\ccsdesc[500]{Information systems~Data analytics}
\ccsdesc[300]{Applied computing~Imaging}
\ccsdesc[300]{Computing methodologies~Neural networks}

\keywords{fMRI, GNNs, brain graphs, neuroimaging analysis, data-centric AI}


\maketitle

\section{Introduction}

The human brain, as a complex network, can be effectively modeled as a graph to study neural activation patterns using neuroimaging. Functional Magnetic Resonance Imaging (fMRI) captures blood-oxygen-level-dependent (BOLD) signals over time, allowing analysis of brain activity via functional connectivity (FC). In these graphs, nodes correspond to regions of interest (ROIs) from a brain parcellation, and edges represent co-activation relationships~\cite{smith2013functional}. This formulation has led to widespread adoption of graph-based approaches in neuroimaging~\cite{luo2024graph, du2024survey}.

Graph machine learning has shown strong promise in analyzing brain graphs for tasks such as disease prediction, cognitive state decoding, and biomarker discovery~\cite{wu2020comprehensive, kim2021learning, said2023neurograph, gadgil2020spatio, dahan2021improving, li2021braingnn, sivgin2024plug}. However, most prior work focuses on model architectures and training objectives, while the upstream graph construction process---from raw fMRI signals to brain networks---remains underexplored within the graph machine learning community. These graphs are typically built using a fixed pipeline based on instantaneous correlations (e.g., Pearson), with limited attention to how preprocessing, connectivity definitions, or feature choices affect downstream performance.

Recent studies have shown that even modest variations in graph construction—such as incorporating dynamic correlations—can significantly impact predictive accuracy~\cite{sivgin2024plug}. This motivates a shift from model-centric innovation toward systematic study of data-centric decisions. In line with principles of Data-Centric AI~\cite{zha2025data,jakubik2024data,jarrahi2022principles}, we argue that the structure and quality of input graphs deserve closer scrutiny. This view is supported by broader AI research showing that improving data representations can yield better performance and faster development than tuning models alone~\cite{ng2025LandingAI}. In graph-based neuroimaging, data-centric learning emphasizes how choices in signal processing, connectivity, and feature design—made during graph construction from raw fMRI—directly influence downstream model performance~\cite{yang2023data}.

\begin{figure*}[t] 
    \centering
    \includegraphics[width=1\linewidth]{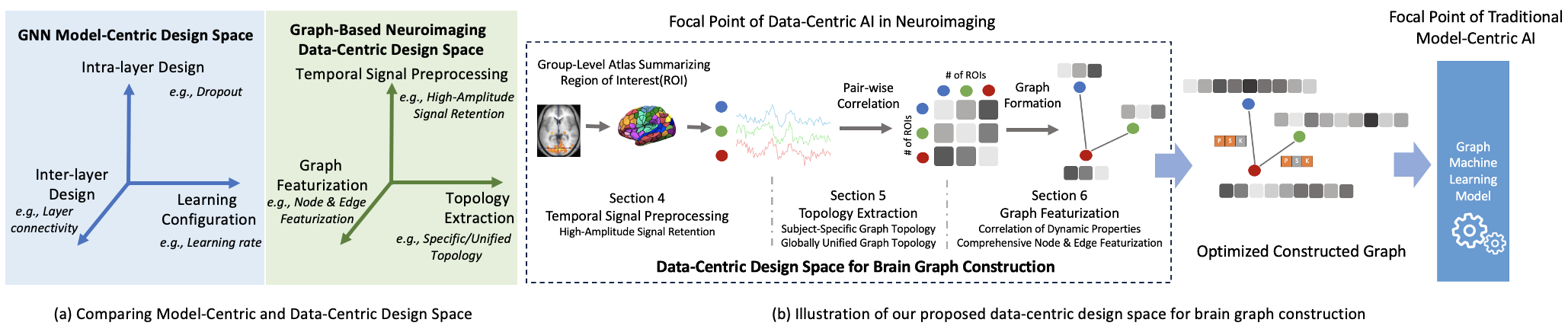}
    \vspace{-4ex}
    \caption{(a) Comparison between model-centric and data-centric design spaces. Traditional approaches focus on model configurations, while our work emphasizes upstream data decisions.
(b) Overview of our proposed data-centric design space for brain graph construction. Instead of relying on a fixed pipeline, we systematically explore signal preprocessing, topology extraction, and graph featurization choices, transforming raw fMRI signals into optimized graph representations for GNNs.}
    \vspace{-2ex}
    \label{fig:pipeline_img}
\end{figure*}

In this work, we reframe brain graph construction as the exploration of a structured \textit{data-centric design space}, which organizes key decisions across three stages: (1) temporal signal processing, (2) topology extraction, and (3) graph featurization. As illustrated in Figure~\ref{fig:pipeline_img}, this design space provides a flexible and structured alternative to the standard fixed pipeline, enabling systematic evaluation of data-centric choices between raw fMRI signals and graph learning models. Thus, rather than proposing new model architectures, we systematically benchmark individual and combined data-centric configurations of existing/modified techniques to assess their impact on neuroimaging graph classification.

We conduct extensive experiments on the HCP1200 and ABIDE datasets and show that thoughtful design choices—such as filtering high-amplitude BOLD fluctuations, applying alternative correlation metrics, unifying connectivity topologies, and encoding lagged dynamics—consistently improve classification performance over standard pipelines. These findings highlight the role of upstream data decisions in shaping brain graph representations. Our benchmark and decomposition of brain graph construction offer a practical toolkit and conceptual foundation for future research in Auto-Data-Centric AI—where data construction pipelines, like model architectures, can be automatically explored and optimized.

Our contributions, framed around the proposed data-centric design space, advance understanding of how upstream data decisions influence graph-based neuroimaging:

\begin{itemize}[left=0pt]
    \item We define and benchmark a data-centric design space for brain graph construction, organized into three stages: temporal signal processing, topology extraction, and graph featurization. This framework highlights how upstream data-centric choices shape brain graphs for downstream learning.
    \vspace{0.25ex}

    \item For temporal signal processing, we evaluate multiple high-amplitude signal retention strategies to assess their impact on co-activation patterns and connectivity formation.
     \vspace{0.25ex}

    \item For topology extraction, we explore both subject-specific and globally unified topology frameworks, and systematically compare correlation metrics (e.g., Pearson, Kendall) for their robustness to outliers and nonlinear dependencies.
     \vspace{0.25ex}

  \item For graph featurization, we incorporate not only static correlations but also dynamic lagged correlations into node features, and explore edge feature encodings to consolidate multiple functional interactions within a single graph representation.

\end{itemize}

The rest of the paper is organized as follows. Section 2 reviews related work. Section 3 presents preliminaries, including notations and datasets. Sections 4–6 explore the data-centric design space across temporal signal processing, topology extraction, and graph featurization. Section 7 presents results, and Section 8 concludes.

\section{Related Work}

\subsection{Deep Learning for Neuroimaging}

Deep learning’s ability to learn highly complex and non-linear patterns from neuroimaging data has enabled learning individualized representations with more predictive power~\cite{arbabshirani2017single, plis2014deep}. Given its ability to detect abstract and complex patterns, DL has been applied in neuroimaging studies of psychiatric and neurological disorders, which are characterized by subtle and diffuse alterations~\cite{vieira2017using}. Compared to standard machine learning methods, deep learning can learn more representative information from the data~\cite{lecun2015deep} to model the brain activity. Despite the advancements in deep learning for neuroimaging, these models heavily depend on the quality and structure of input data~\cite{yan2022deep}. This motivates our study to investigate data-centric AI approaches for optimizing neuroimaging data for improving downstream prediction performance.

\subsection{Data-Centric AI on Graphs}
Data-Centric AI focuses on systematically diagnosing and addressing real-world data issues through AI-driven methods~\cite{zha2023data}. While AI research has primarily advanced model development, the growing demand for Data-Centric AI~\cite{singh2023systematic} arises from the fact that even state-of-the-art models underperform on noisy, uncurated data—illustrated by the "garbage in, garbage out" phenomenon.

Graph data, which models entities and relationships, requires a distinct approach due to its manually abstracted nature. Research in Data-Centric Graph Learning has explored how graph data and models interact~\cite{yang2023data, wang2022improving}. Given graph data's unique characteristics, we can analyze different stages of graph learning through a Data-Centric AI lens. This involves addressing key questions: how to construct, when and where to modify, and how to refine a graph. In this work, we adapt the principles of data-centric graph learning to the neuroimaging domain by analyzing how temporal 3D signals preprocessing, topology extraction, and graph featurization affect the construction of functional brain graphs from fMRI data.

\subsection{Model-Centric Design Space for Graph Machine Learning}

The high-impact design space paradigm introduced by GraphGym~\cite{you2020design} enhances GNN development by systematically exploring architectural hyperparameters to identify optimal configurations across diverse tasks. Their work establishes a principled approach to model-centric optimization, ensuring generalizable performance. Inspired by this framework, we extend this paradigm into the realm of data-centric AI. Instead of tuning model hyperparameters, we conceptualize data-centric ``hyperparameters''—a structured set of preprocessing strategies and feature engineering choices that critically shape model performance. By integrating the most effective proposed methods, we construct a generalizable and empirically robust data-centric workflow, optimizing the entire pipeline of deep learning in neuroimaging rather than just the model architecture. By leveraging this data-centric design space idea to evaluate the integration of our proposed methods, the overall enhancement compared to baseline is shown in Figure ~\ref{fig:finaltable}.

\section{Preliminaries} 

Here we first introduce the general structure of preprocessing raw fMRI data into graphs, and then provide a formal definition. \vspace{1ex} 

\textbf{Standard fMRI to Graph Construction.}
Constructing brain graphs from fMRI data typically follows a fixed pipeline that transforms voxel-level signals into region-level functional connectivity graphs. This pipeline involves preprocessing the fMRI signals, summarizing them within regions of interest (ROIs), computing pairwise correlations to obtain a functional connectivity (FC) matrix, and applying thresholding to define graph edges. A more detailed description of the standard pipeline is provided in Appendix~\ref{sec:standard_pipeline}.\vspace{1ex}

\label{sec:preliminaries}

\textbf{Notations and Definitions.}
Here we formally define the functional brain graph created from the standard fMRI to graph pipeline.

\begin{definition}
    The brain functional connectomics graph is defined as \( G = (\mathcal{V}, \mathcal{E}, \mathbf{X}, \mathbf{E}) \), where the node set \( \mathcal{V} = \{v_1, v_2, \ldots, v_n\} \) corresponds to regions of interest (ROIs) defined by a parcellation atlas, and the edge set \( \mathcal{E} \subseteq \mathcal{V} \times \mathcal{V} \) represents functional connectivity (FC) above a defined threshold between pairs of ROIs. The matrix \( \mathbf{X} \in \mathbb{R}^{n \times d} \) denotes the node feature matrix, where \( n \) is the number of ROIs and \( d \) is the feature dimension. The adjacency matrix \( \mathbf{A} \in \{0, 1\}^{n \times n} \) encodes the graph structure, with \( \mathbf{A}_{ij} = 1 \) indicating an edge between nodes \( v_i \) and \( v_j \), signifying a strong co-activation relationship between the corresponding ROIs. The edge feature matrix \( \mathbf{E} \in \mathbb{R}^{|\mathcal{E}| \times d'} \) encodes additional multi-view connectivity information across multiple correlation measures. 
\end{definition}

Note that a more comprehensive summary of notations can be found in Table~\ref{tab:notations} in Appendix~\ref{sec:notations}.

\section{Temporal Signal Preprocessing}
\label{sec:denoising}

fMRI is a powerful non-invasive tool for studying brain function, widely used in Neuroscience, Psychology, and Psychiatry~\cite{smith2004overview}. However, fMRI data, derived from the BOLD signal, may be influenced by non-neural sources of variability~\cite{power2017sources}, necessitating careful preprocessing to optimize the quality of extracted features. Numerous techniques are available for denoising fMRI data~\cite{caballero2017methods}, which are crucial for further analysis in neuroimaging. For brain network analysis using fMRI data, such extensive preprocessing is essential before any analysis can be conducted~\cite{smith2013functional}. Nevertheless, even after common preprocessing steps, the BOLD signal often remains noisy, which can affect the accuracy of correlation calculations, a key step in graph construction.

\textit{Motivating High-Amplitude Signal Retention.}
We observed that the approach for calculating functional connectivity can be influenced by both high- and low-amplitude fluctuations. Motivated by this, previous neuroimaging studies have also explored related research questions from a domain knowledge perspective. Previous studies on resting-state fMRI data have demonstrated that high-amplitude BOLD fluctuations may play a crucial role in defining brain functional connectivity, as they often correspond to meaningful co-activation patterns. Retaining primarily high-amplitude fluctuations has been shown to reduce data volume by up to 94\%~\cite{tagliazucchi2012criticality}, while still preserving key functional networks~\cite{liu2013time}. Additionally, other work has shown that functional networks are driven by activity at only a few critical time points, using just 15\% of the data to accurately replicate a resting-state functional network~\cite{liu2013time}. To investigate this, we explore several variants of high-amplitude signal retention as a preprocessing design choice. 

\subsection{High-Amplitude Signal Retention Process}
Let $\mathcal{R}$ be the set of ROIs and $\mathcal{T} = \{1,2,...,T\}$ be the set of time points. For each ROI $r \in \mathcal{R}$, we define its temporal BOLD signal as:
\begin{equation}
\small 
\mathbf{T}_r = [T_r(1), T_r(2), ..., T_r(T)] \in \mathbb{R}^T
\end{equation}
where $T_r(t)$ represents the BOLD signal value at time $t$ for ROI $r$.  

Formally, we define the high-amplitude signal retention process as: $f_{\text{high-amplitude}}: \mathbb{R}^T \rightarrow \mathbb{R}^T$, which consists of two steps: normalization and thresholding.

Our first step in normalization is built on the classical Z-score normalization, which is defined as follows:
\begin{equation}
\small 
\mathbf{Z}_r = [Z_r(1), ..., Z_r(T)], \text{ where } Z_r(t) = \frac{T_r(t) - \mu_r}{\sigma_r}
\end{equation}

\begin{equation}\label{eq:stdev}
\small 
\mu_r = \frac{1}{T}\sum_{t=1}^T T_r(t), \quad
\sigma_r = \sqrt{\frac{1}{T}\sum_{t=1}^T (T_r(t) - \mu_r)^2}
\end{equation}
where $\mu_r$ and $\sigma_r$ are the mean and standard deviation of ROI $r$'s signal $\mathbf{T}_r$, respectively.

After normalizing we apply a thresholding function, \( \Theta(t, \mathbf{Z}_r; \theta, \gamma) \), that allows for multiple variants based on the threshold \( \theta \) and a parameter \( \gamma \), as follows:
\vspace{-1ex}
\begin{equation}
\small 
\Theta(t, \mathbf{Z}_r; \theta, \gamma) = 
\begin{cases}
Z_r(t), & \text{if } |Z_r(t)| \geq \theta \text{ and } \gamma = 0 \\
1, & \text{if } |Z_r(t)| \geq \theta \text{ and } \gamma = 1 \\
0, & \text{otherwise}
\end{cases}
\end{equation}

This function allows for two distinct behaviors. When \( \gamma = 0 \), the function retains values greater than or equal to \( \theta \), while setting all values below the threshold to zero. This is referred to as thresholding with retention and allows a continuous representation while emphasizing stronger signals. This setting ensures that high-magnitude fluctuations are preserved without introducing artificial saturation effects. On the other hand, when \( \gamma = 1 \), the function converts values greater than or equal to \( \theta \) to 1, and all values below the threshold to 0 (i.e., a binarization based on \( \theta \)). This variant is called binary thresholding and enforces a strict thresholding mechanism  that highlights only the most significant fluctuations. Notably, $\gamma = 1$ provides a balance between emphasizing high-amplitude signals and mitigating potential outliers, which retains the relative differences in high-amplitude regions while preventing extreme values from disproportionately influencing the connectivity computation.

We explore two common strategies to determine the threshold \( \theta \) can be determined in two ways. The first approach is the percentile-based method, where \( \theta \) is defined as the \( \alpha \)-th percentile of \( \mathbf{Z}_r \), denoted by \( Q_{\alpha}(\mathbf{Z}_r) \), with \( \alpha \in (0, 1) \). The second way is the standard deviation-based method, where \( \theta \) is computed as \( \beta \sigma \), with \( \sigma \) representing the global standard deviation of \( \mathbf{Z}_r \) across all ROI $r$ and \( \beta > 0 \) as a scaling factor. 

\begin{figure}[t] 
    
    \includegraphics[width=0.9\linewidth]{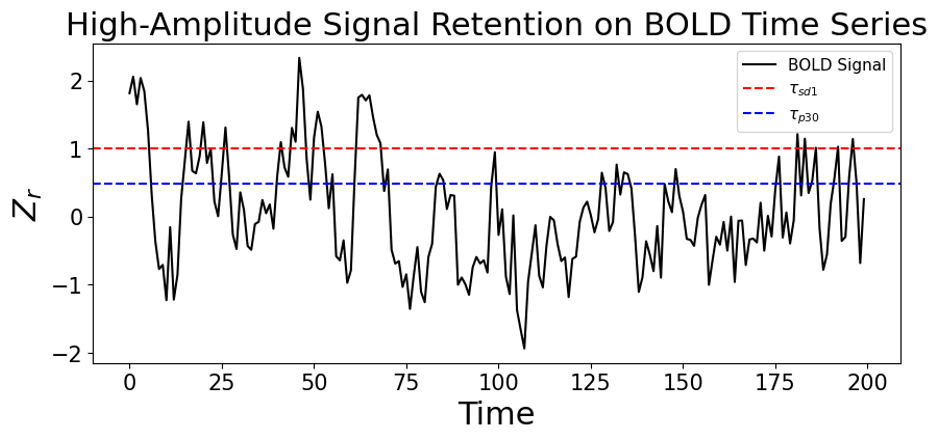}
    \vspace{-2.5ex}
    \caption{High-Amplitude Signal Retention Process.}
    \vspace{-3ex}
    \label{fig:denoising_img}
\end{figure}

The High-Amplitude Signal Retention process is visualized in Figure~\ref{fig:denoising_img}. In summary, this flexible thresholding framework allows for both continuous and binary transformations of the data, depending on the desired outcome. In this work, we fixed $\alpha=30$ and $\beta=1$ (i.e., top 30 percentile and 1 standard deviation, respectively). Thus, we can summarize our four high-amplitude signal retention variants as  $\Theta_{\theta_{p30}\gamma_1}$, $\Theta_{\theta_{p30}\gamma_0}$, $\Theta_{\theta_{sd1}\gamma_0}$, $\Theta_{\theta_{sd1}\gamma_1}$.

\section{Topology Extraction}

Given our Data-Centric AI perspective, we treat the topology of the brain graph—not as a fixed byproduct of fMRI processing—but as a tunable design choice. While traditional pipelines extract topology from a correlation matrix using default heuristics, we investigate how alternative design decisions can shape the connectivity structure to better support graph-based learning. In this section, we explore two complementary directions: (1) modifying the correlation-based construction of subject-specific topologies, and (2) exploring a globally unified graph topology shared across subjects. These approaches reflect different points in the design space and highlight how edge-level decisions impact downstream performance.

After extracting the signal data and averaging the time courses of ROIs based on a predefined atlas, correlation analysis is the most widely used approach for functional connectivity construction/analysis (i.e., topology extraction)~\cite{du2024survey}. This allows for the simple and direct computation of a subject-specific \( n \times n \) correlation matrix~\cite{smith2013functional}. 

However, a fixed correlation method may not always be optimal, and vary in ability to handle low-amplitude signal fluctuations~\cite{smith2013functional} or oversimplify the complex patterns of brain activity~\cite{ahmadi2023comparative}. Furthermore, subject-specific brain graph structures exhibit variability, as edges are determined individually for each subject. 
This motivates exploration of alternative topological designs within a data-centric framework.

\subsection{Subject-Specific Brain Graph Topology} 
\label{sec:spearman_kendall}

 Here for each individual subject-specific brain we explore alternative correlation measures as design choices for defining functional connectivity, evaluating their potential to offer complementary perspectives in extracting topology for functional brain connectivity graphs.

\subsubsection{Motivation}
Pearson Correlation Coefficient (PCC), while widely used in functional connectivity analysis for its simplicity, has notable limitations. It assumes linear relationships between time series, which may not capture the complex dependencies in brain activity~\cite{ahmadi2023comparative}. Moreover, PCC's lack of robustness means outliers can introduce false correlations or mask existing ones~\cite{wilcox2023updated}. When dealing with non-normal distributions and outliers, researchers have suggested alternative measures like Spearman's rank correlation and Kendall's tau~\cite{rousselet2012improving}. These metrics may offer advantages in handling non-linear dependencies and are shown to improve predictive performance ~\cite{ahmadi2023comparative}.

\subsubsection{Correlation-based Edge Construction} Based on the above, we first formally define the traditionally used Pearson Correlation Coefficient, and then also Spearman Correlation and Kendall's tau that we propose exploring to investigate their potential as being more robust (e.g., against outliers). The detailed definitions of three correlation coefficient can be found in Appendix~\ref{sec:correlation}

We examined subject-specific topology through different correlation measures as design choices. While Pearson is most widely used, Spearman and Kendall offer robustness to nonlinearity and outliers, highlighting how simple substitutions in edge construction can affect brain graph quality.

\subsection{Globally Unified Brain Graph Topology}
\label{sec:unified_graph}
The typical approach in brain graph construction is to define topology separately for each subject by thresholding their correlation matrix to retain only the strongest functional connections. This results in individualized graphs that vary across subjects. While widely adopted, the implications of such subject-specific variability for GNN-based learning—especially from a data-centric perspective—remain insufficiently understood.

\subsubsection{Motivation.} 
From a data-centric perspective, subject-specific brain graphs may underutilize GNN aggregation, as each node already encodes global connectivity through its features. Unlike typical graph tasks where neighborhoods carry new information, brain graphs derived from fMRI embed full co-activation profiles at the node level. While prior work has explored learning-based strategies to adapt graph structure during training~\cite{cui2022interpretable, li2023interpretable}, we instead consider a simple, data-driven alternative: constructing a shared, stable topology by identifying edges that consistently appear across subjects.

\subsubsection{Global Functional Co-activation Filtering.} 
\begin{figure}[t] 

    \includegraphics[width=0.85\linewidth]{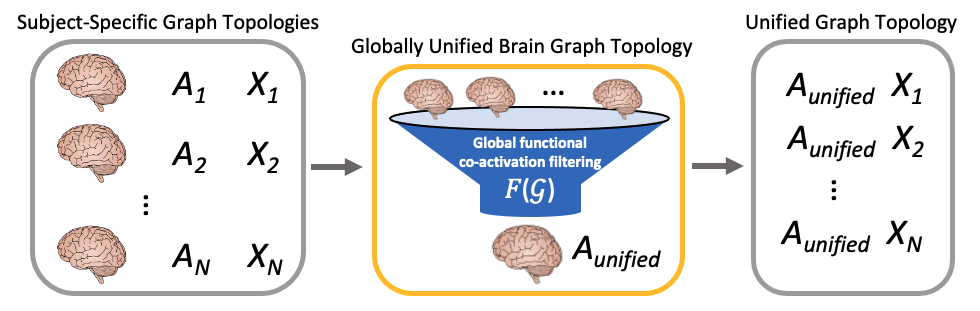}
    \vspace{-1.75ex}
    \caption{Globally unified brain graph topology process. }
    \vspace{-2.5ex}
    \label{fig:unified_graph}
\end{figure}

We explore a data-driven design choice for constructing a globally consistent graph topology. 
This configuration assigns a shared graph topology to all subjects in the dataset by assigning a consistent adjacency matrix. The high-level idea of Global Functional Co-activation Filtering is visualized in Figure ~\ref{fig:unified_graph} 
Specifically, we define the set of all brain graphs in the dataset as: $\mathcal{G} = \{ G_1, G_2, \dots, G_N \}.$
Then, to construct a globally unified topology, we define a function \( F \) that aggregates the adjacency matrices of all subject-specific brain graphs and outputs a unified adjacency matrix: $F(\mathcal{G}) \rightarrow \mathbf{A}_{\text{unified}}.$

The entries of \( \mathbf{A}_{\text{unified}} \) are determined by retaining the most frequently occurring edges across subjects. Specifically, we set a threshold \( k_{\text{threshold}} \) to include only the top proportion (e.g., 5\%) of the most consistently present connections:
\begin{equation}
{A}_{\text{unified}, ij} =
\begin{cases} 
1, & \text{if } \sum_{i=1}^{N} {A}_{i, ij} \geq k_{\text{threshold}} \\
0, & \text{otherwise}
\end{cases}
\end{equation}
where \( k_{\text{threshold}} \) is chosen such that the total number of edges in \( \mathbf{A}_{\text{unified}} \) does not exceed the desired proportion of possible connections.
Thus, the final unified graph is: $G_{\text{unified}} = (\mathcal{V}, \mathbf{A}_{\text{unified}}, \mathbf{X}),$
\noindent where \( \mathbf{X} \) represents the node features of individual subjects, and all subjects share the same unified adjacency matrix \( \mathbf{A}_{\text{unified}} \).

This design choice emphasizes broadly co-activated regions to reduce variability and promote a more stable functional connectivity for GNNs. Consequently, it supports a more meaningful aggregation mechanism in GNNs by focusing on generalized connectivity patterns. While graph structure learning has shown promise in modeling individualized connectivity patterns~\cite{cui2022interpretable,li2023interpretable}, our findings suggest that even simple group-level topologies can provide strong performance baselines.

\section{Advanced Graph Featurization}

Featurization determines how functional connectivity is encoded in brain graphs and directly impacts what information GNNs can leverage. While prior work often treats static and dynamic functional connectivity separately, we view them as complementary design choices in a unified featurization space. We explore incorporating lagged temporal correlations into node features and propose edge feature encodings that capture multi-view functional relationships beyond the static Pearson graph.

\subsection{Correlation of Dynamic Properties}
\label{sec:lag}

The common approach to calculating functional connectivity (FC) relies on computing correlations between the time series of each pair of nodes~\cite{said2023neurograph, cui2022braingb, li2021braingnn, luo2024graph}. As discussed in Section \ref{sec:spearman_kendall}, previous studies typically use this \(n \times n\) correlation matrix as the node features, representing the connectivity between a specific ROI and all other ROIs. While this approach is straightforward, using only the instantaneous (i.e., zero-lag) correlation between inter-regional time courses may fail to capture important brain dynamics such as causality and directionality~\cite{marrelec2006partial}, prompting researchers to explore dynamic properties of fMRI data. Recent works have focused on developing advanced GNNs to model dynamic brain graphs~\cite{kim2021learning, gadgil2020spatio, zhu2024spatio}. Although advanced models enhance the ability of GNNs to capture temporal properties, they also introduce increased complexity, and can sometimes degrade performance~\cite{said2023neurograph}. Here we explore a data-centric design choice that incorporates lagged correlations into brain graph construction, aiming to improve sensitivity to temporal dependencies in fMRI signals.

From a data-centric perspective, the common approach of calculating only the instantaneous correlation may fail to capture crucial temporal dynamics, overlooking time-lagged interactions between brain regions. For example, two ROIs may show strong overall correlation, but if the activation of one lags behind the other, PCC would inaccurately suggest a weak connection, missing valuable information about neural processes. By incorporating dynamic properties, we can enable brain graphs to better capture lagged interactions and the potential temporal propagation of neural signals.

From a neuroimaging perspective, researchers have been investigated the casualty and directionality in the neural activity of human brain for a long time~\cite{smith2013functional, du2024survey, smith2011network}. 
Previous studies have demonstrated that dynamic inter-regional interactions can be captured by lagged correlation~\cite{sivgin2024plug}. 
Also, researchers have investigated the idea of Cross-Correlation~\cite{fiecas2013quantifying, hyde2012cross}, formalized as: 
\begin{equation}
\small 
R_{r_i, r_j}(\delta) = \frac{\sum_{t} (\mathbf{T}_{r_i}(t) - \mu_{r_i})(\mathbf{T}_{r_j}(t + \delta) - \mu_{r_j})}
  {\sqrt{\sum_{t} (\mathbf{T}_{r_i}(t) - \mu_{r_i})^2} \sqrt{\sum_{t} (\mathbf{T}_{r_j}(t + \delta) - \mu_{r_j})^2}}
  \label{eq:crosscorrelation}
\end{equation}
\noindent where \( \mathbf{T}_{r_i}(t) \) and \( \mathbf{T}_{r_j}(t + \delta) \) represent the BOLD signal values at time \( t \) and \( t + \delta \) for ROIs \( r_i \) and \( r_j \), respectively, while \( \mu_{r_i} \) and \( \mu_{r_j} \) are the mean values of the BOLD signals of ROI \( r_i \) and ROI \( r_j \).

\begin{figure}[t] 
    \hspace{\linewidth}
    \includegraphics[width=0.95\linewidth]{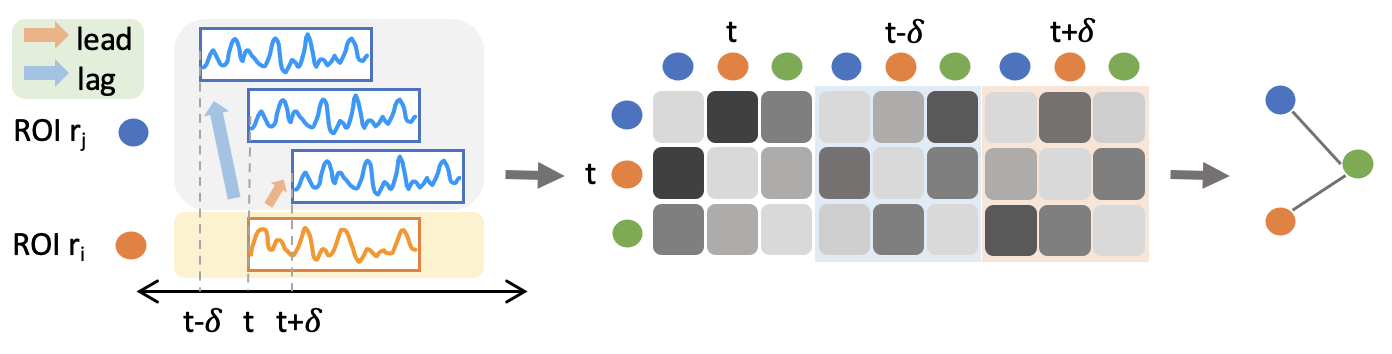}
    \vspace{-1ex}
    \caption{An illustration of lagged correlation computation. } 
    \label{fig:lag_img}
    \vspace{-1.5ex}
\end{figure}

As illustrated in Figure~\ref{fig:lag_img}, we explore constructing brain graphs with lagged correlations as a design choice to capture leading/lagging relationships between regions. Specifically, we shift the time series of each ROI relative to the others by certain time points and then calculate the leading and lagging relations. As shown in Equation~\ref{eq:crosscorrelation}, if \(\delta\) is positive, it means that \( \mathbf{T}_{r_i}(t) \) leads \( \mathbf{T}_{r_j}(t) \), indicating that changes in ROI \( r_i \) precede those in ROI \( r_j \). Similarly, if \(\delta\) is negative, it means \( \mathbf{T}_{r_i}(t) \) lags \( \mathbf{T}_{r_j}(t) \),
i.e., changes in ROI \( r_i \) follow those in \( r_j \). This is repeated for all ROI pairs in the graph.

\subsection{Comprehensive Node \& Edge Featurization}
\label{sec:edgefeature}

The previous sections introduced multiple design dimensions for brain graph construction, each targeting different aspects of the data. Here, we investigate how combining these design choices—on signal selection, correlation measures, and temporal sensitivity—can improve node and edge featurization. These strategies represent different axes in the data-centric design space, and combining them allows us to evaluate richer representations through expanded node and edge features. Since each of these strategies targets a different stage of brain network construction, here we can combine them to investigate a more comprehensive node/edge featurization. More specifically, we define the node and edges features as follows: 

\vspace{0.5ex}
\noindent \textit{Node features} are enriched by concatenating correlation matrices. 

\noindent Specifically, we construct the new node feature matrix as:
 $\mathbf{X} = \big[ \mathbf{R}^{(1)}, \mathbf{R}^{(2)}, \dots, \mathbf{R}^{(d')} \big] \in \mathbb{R}^{n \times d'n}$, 
where \( \mathbf{R}^{(i)} \) represents the correlation matrix computed using the \( i \)-th correlation measure, including Pearson, Spearman, Kendall, and lagged correlations.

\vspace{0.5ex}
\noindent \textit{Edge features} are typically not included in standard brain graph representations, or perhaps sometimes weights according to the correlation values. However, here we investigate whether encoding the multiple correlations as edge features can provide benefits beyond their contribution as node features. Specifically, here we retain the adjacency matrix from the Pearson correlation graph, which serves as the primary graph structure and then construct an edge feature matrix as follows:
$\mathbf{E} \in \{0,1\}^{|E| \times d'}, ~\mathbf{E}_{(i,j),k} = \mathbb{I} \big( \mathbf{R}^{(k)}_{ij} > \rho \big)$, with $d'$ correlation measures and $\rho$ representing the correlation threshold to determine if edge (i,j) in correlation matrix  \( \mathbf{R}^{(k)} \) is encoded as 1/0 in $\mathbf{E}_{(i,j),k}$ according to the indicator function, $\mathbb{I}$.

\section{Experiments}

In this section, we systematically evaluate how different data-centric design choices affect downstream GNN performance on neuroimaging tasks. We conduct controlled experiments across our three dimensions---temporal signal processing, topology extraction, and graph featurization---to assess their individual and combined impact. Our findings demonstrate that structured exploration of this design space yields significant performance gains over standard fixed-pipeline approaches.

\subsection{Experiment Settings}
We implemented a unified GNN architecture across all datasets to maintain a consistent data-centric approach. For the HCP dataset, we used a three-layer GNN, while for the ABIDE dataset, we used a two-layer GNN, and both used sort pooling~\cite{zhang2018end} followed by a two-layer Multi-Layer Perceptron (MLP)~\cite{said2023neurograph}.

Note that we randomly split the data into 70\% training, 20\% testing, and 10\% validation, 
with stratified partitioning to ensure equitable class distribution across splits.
Each model was trained for 100 epochs with a learning rate of 0.0005 for classification tasks. Throughout all experiments, we applied a dropout rate of 0.5, weight decay of 0.0005, and used 32 hidden dimensions for the GNN convolution layers and 64 hidden dimensions for the MLP layers. Cross-entropy was used as the loss function for all classification tasks. 

\textbf{Datasets.}
 Human Connectome Project (HCP) S1200~\cite{van2013wu} and Autism Brain Imaging Data Exchange (ABIDE) dataset~\cite{di2014autism}. For HCP, we use both resting-state (1,078 subjects) and task-based fMRI scans (>7,000 scans across 7 task types), framing the problems as gender classification (2-class) and task-type classification (7-class), respectively. For ABIDE, we use resting-state scans from 1,009 subjects with binary labels indicating autism diagnosis. All datasets were parcellated using the Schaefer atlas and underwent standard preprocessing pipelines; full details are provided in Appendix~\ref{sec:appendix_dataset}.

\begin{figure}[t]
    \centering
    \includegraphics[width=0.9\linewidth]{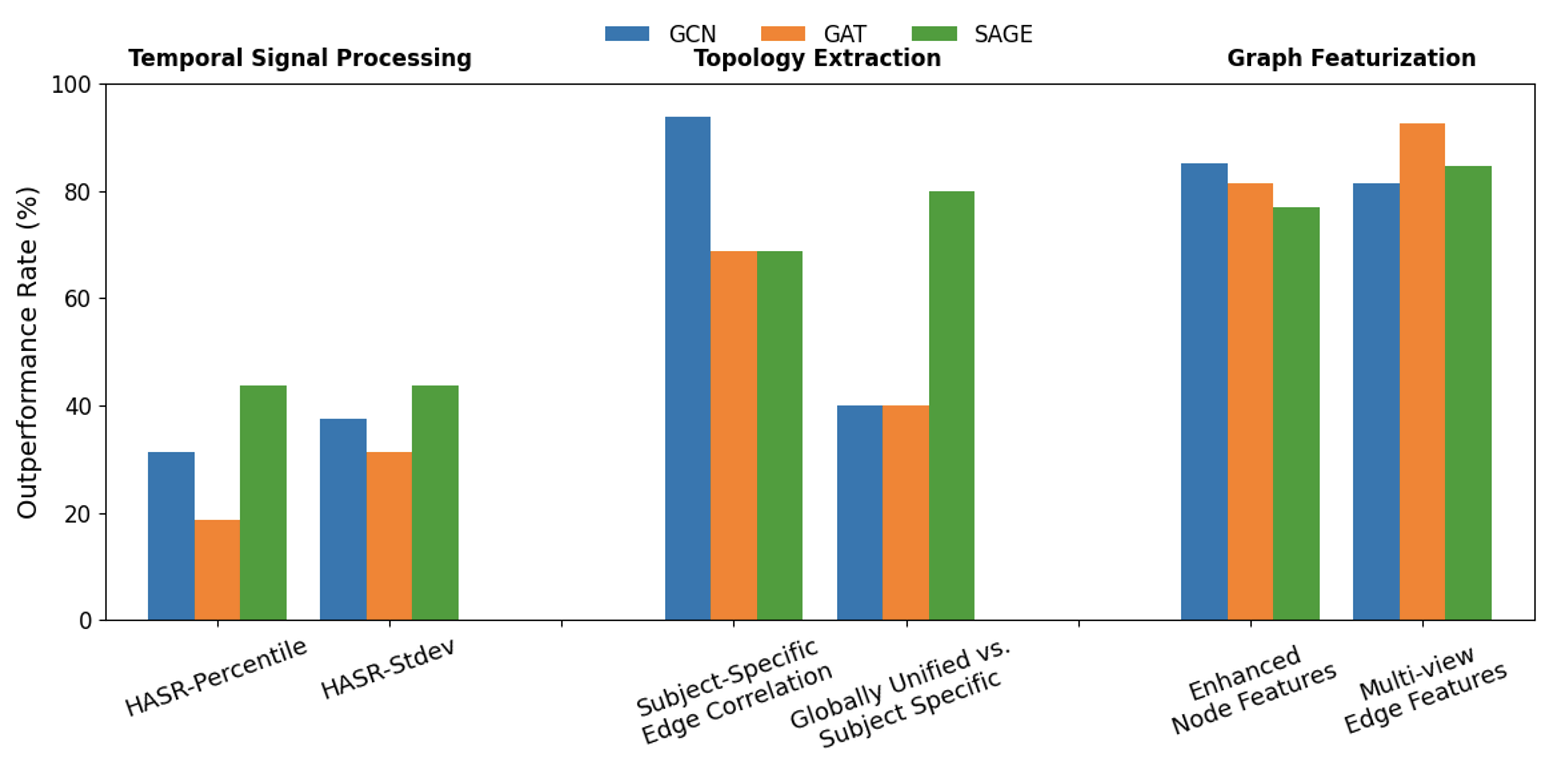}
    \vskip -2.5ex
    \caption{Outperformance rate of each data-centric design choice across 8 settings (covering the ROIs and 2 datasets). Higher values indicate more consistent improvements over the baseline, i.e., not leveraging the data-centric design space.}
    \vskip -3ex
    \label{fig:outperform}
\end{figure}

\subsection{Overall Results: Data-Centric Design Space}

In this section, we synthesize empirical findings from our exploration of the data-centric design space, assessing which design dimensions—temporal signal preprocessing, topology extraction, and graph featurization—consistently improve performance across datasets, models, and ROI settings. Next, we provide a summary of the key findings from these results: 

\begin{figure}[t] 
    \includegraphics[width=\linewidth]{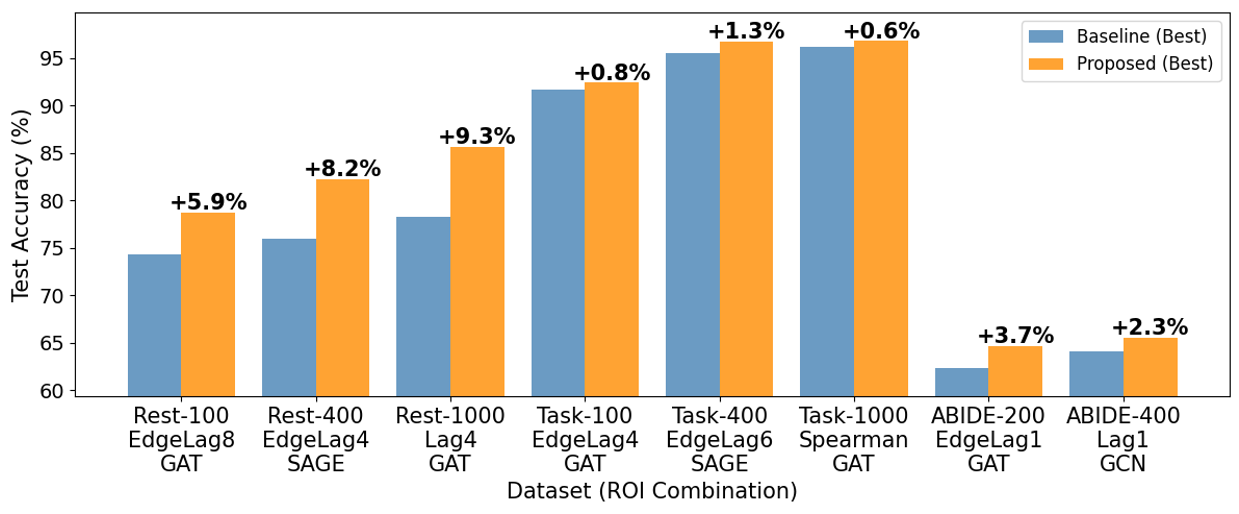}
    \vskip -2ex
    \caption{
    A comparison of the best baseline model using the fixed standard fMRI to graph construction pipeline (blue) as compared to systematic exploration within a data-centric design space (orange). We observe consistent improvement with the inclusion of not only model hyperparameters, but graph construction hyperparameters.
    }
    \label{fig:finaltable}
    \vskip -2ex
\end{figure}

\begin{itemize}[left=0pt]

    \item Table~\ref{tab:ranking} compares the performances of each individual proposed component. We observe incorporating lagged correlation and advanced featurization generally achieve top ranks.

    \item Figure~\ref{fig:outperform} provides an overall perspective of the benefits of our data-centric design space. Specifically, we present how often the different dimensions are able to outperform the baseline, i.e., just using the standard fixed fMRI to brain graph pipeline. These results are aggregated over all dataset and ROI settings. This experiment captures how consistently each design leverages data to gain improvements, offering additional insight into the reliability of specific strategies across contexts.
    
    \item As shown in Figure~\ref{fig:finaltable}, our approach consistently outperforms the baseline across datasets by selecting the best-performing GNN based on validation performance and evaluating on the test set. For HCP-Task, although the baseline already achieves high performance, the proposed data-centric strategies further enhance the results, confirming their complementary value. The overall improvements demonstrate the robustness and generalizability of our data-centric design space exploration.

    \item Lastly, we observe that no single strategy dominates across all datasets. This reinforces the value of a modular and flexible data-centric design space framework that enables systematic comparison and selection.
\end{itemize}

\begin{table*}[t]
\centering
\scriptsize
\caption{Average accuracy ranking (across GNNs and ROIs) of representative data-centric design space components.}
\vskip -2.75ex
\label{tab:ranking}
\resizebox{\textwidth}{!}{%
\begin{tabular}{lcccccccccccc}
\toprule
 & & \multicolumn{4}{c}{\textbf{Temporal Signal Processing}} & \multicolumn{2}{c}{\textbf{Topology Extraction}} & \multicolumn{5}{c}{\textbf{Graph Featurization}} \\
\cmidrule(lr){3-6} \cmidrule(lr){7-8} \cmidrule(lr){9-13} 
 & \textbf{Baseline} & \textbf{\(\Theta_{\theta_{p30}\gamma_0}\)} & \textbf{\(\Theta_{\theta_{p30}\gamma_1}\)} & \textbf{\(\Theta_{\theta_{sd1}\gamma_0}\)} & \textbf{\(\Theta_{\theta_{sd1}\gamma_1}\)} & \textbf{Spearman} & \textbf{Kendall} & \textbf{Lag = 1} & \textbf{Lag = 5} & \textbf{w/ Edge Feat.} & \textbf{E.F.Lag = 1} & \textbf{E.F.Lag = 5} \\
\midrule
\textbf{Rest}  & 8.7  & 7.0  & 5.1  & 8.9  & 4.7  & 8.2  & 4.4  & 10.1 & 2.5  & 9.1  & 5.3  & 2.3  \\
\textbf{Task}  & 5.9  & 8.2  & 6.9  & 10.1 & 9.6  & 4.0  & 3.7  & 4.8  & 1.3  & 5.2  & 4.8  & 2.3  \\
\textbf{Abide} & 4.8  & 6.8  & 7.8  & 7.0  & 5.0  & 4.8  & 4.3  & 3.5  & 4.7  & 6.5  & 10.3 & 11.8 \\
\bottomrule
\end{tabular}%
}
\vskip -1.5ex
\end{table*}
\begin{table*}[ht]
\centering
\caption{Classification results on HCP-Rest and HCP-Task for various high-amplitude signal retention strategies.}
\vskip -2.5ex
\label{tab:denoising_rs}

\resizebox{\textwidth}{!}{%
\begin{tabular}{clccccccccccccc}
\toprule
& & \multicolumn{4}{c}{\textbf{ROI 100}} & \multicolumn{4}{c}{\textbf{ROI 400}} & \multicolumn{4}{c}{\textbf{ROI 1000}} & \textbf{Avg. across}  \\

\cmidrule(lr){3-6} \cmidrule(lr){7-10} \cmidrule(lr){11-14}
\textbf{Dataset} &  & \textbf{GCN} & \textbf{GAT} & \textbf{SAGE} & \textbf{Avg. across GNNs} & \textbf{GCN} & \textbf{GAT} & \textbf{SAGE} & \textbf{Avg. across GNNs} & \textbf{GCN} & \textbf{GAT} & \textbf{SAGE} & \textbf{Avg. across GNNs} & \textbf{ROIs \& GNNs} \\
\midrule
\multirow{5}{*}{\rotatebox{90}{\large \textbf{HCP-Rest~~}}} 
& \textbf{Baseline} & 74.07 ± 1.13 & 74.35 ± 2.56 & 71.11 ± 3.76 & 73.18 & 74.63 ± 0.68 & 76.02 ± 0.80 & 75.37 ± 2.08 & 75.34 & 81.39 ± 2.81 & 78.33 ± 3.43 & 81.48 ± 1.80 & 80.40 & 76.31 \\
& \textbf{\(\Theta_{\theta_{p30}\gamma_0}\)} & 71.39 ± 3.05 & 73.15 ± 2.65 & 74.44 ± 1.35 & 72.99 & 77.41 ± 2.51 & 75.56 ± 2.04 & 76.76 ± 1.42 & 76.58 & 82.78 ± 0.68 & 81.85 ± 2.75 & 83.61 ± 1.12 & 82.75 & 77.44 \\
& \textbf{\(\Theta_{\theta_{p30}\gamma_1}\)} & 72.31 ± 2.74 & 73.52 ± 1.74 & 71.02 ± 1.39 & 72.28 & 78.06 ± 1.98 & 76.76 ± 3.15 & 77.96 ± 1.19 & 77.59 & 84.35 ± 1.29 & 84.63 ± 1.72 & 85.09 ± 0.80 & 84.69 & 78.19 \\
& \textbf{\(\Theta_{\theta_{sd1}\gamma_0}\)} & 69.91 ± 1.01 & 72.41 ± 1.62 & 72.13 ± 1.81 & 71.48 & 76.30 ± 1.15 & 75.37 ± 1.72 & 76.57 ± 2.35 & 76.08 & 81.39 ± 1.07 & 82.96 ± 1.53 & 81.85 ± 2.18 & 82.07 & 76.54 \\
& \textbf{\(\Theta_{\theta_{sd1}\gamma_1}\)} & 74.26 ± 1.59 & 71.67 ± 3.48 & 72.22 ± 3.70 & 72.72 & 78.43 ± 2.76 & 77.22 ± 1.50 & 77.22 ± 2.44 & 77.62 & 84.44 ± 1.77 & 84.07 ± 1.26 & 83.89 ± 1.45 & 84.13 & 78.16 \\
\bottomrule

\multirow{5}{*}{\rotatebox{90}{\large \textbf{HCP-Task~~}}} 
& \textbf{Baseline} & 91.05 ± 0.29 & 91.93 ± 0.50 & 91.71 ± 0.30 & 91.56 & 94.57 ± 0.60 & 95.08 ± 0.51 & 95.51 ± 0.19 & 95.05 & 96.06 ± 0.36 & 96.26 ± 0.62 & 96.49 ± 0.28 & 96.27 & 94.30 \\
& \textbf{\(\Theta_{\theta_{p30}\gamma_0}\)} & 89.78 ± 0.50 & 90.36 ± 0.64 & 90.07 ± 0.91 & 90.07 & 93.35 ± 0.39 & 93.51 ± 0.43 & 94.67 ± 0.35 & 93.84 & 95.08 ± 0.48 & 95.71 ± 0.66 & 96.84 ± 0.17 & 95.88 & 93.26 \\
& \textbf{\(\Theta_{\theta_{p30}\gamma_1}\)} & 90.91 ± 0.34 & 90.85 ± 0.37 & 90.79 ± 0.64 & 90.85 & 93.51 ± 0.21 & 94.01 ± 0.49 & 95.33 ± 0.34 & 94.28 & 95.27 ± 0.14 & 96.14 ± 0.62 & 97.18 ± 0.30 & 96.20 & 93.78 \\
& \textbf{\(\Theta_{\theta_{sd1}\gamma_0}\)} & 87.62 ± 0.85 & 87.79 ± 0.56 & 87.62 ± 0.67 & 87.68 & 91.47 ± 0.82 & 92.87 ± 0.68 & 93.57 ± 0.42 & 92.64 & 94.03 ± 0.39 & 95.14 ± 0.27 & 96.32 ± 0.84 & 95.16 & 91.83 \\
& \textbf{\(\Theta_{\theta_{sd1}\gamma_1}\)} & 88.03 ± 0.93 & 88.69 ± 0.47 & 88.60 ± 0.85 & 88.44 & 92.76 ± 0.78 & 93.49 ± 0.60 & 93.70 ± 0.38 & 93.32 & 93.59 ± 0.85 & 95.41 ± 0.34 & 96.30 ± 1.50 & 95.10 & 92.29 \\
\bottomrule

\end{tabular}%
}
\vskip -1.5ex
\end{table*}

\begin{table*}[h]
\centering
\caption{Classification results on HCP-Rest and HCP-Task datasets under different correlation strategies and ROI settings.}
\vskip -2.25ex
\label{tab:corr_merged}
\resizebox{\textwidth}{!}{%
\begin{tabular}{clccccccccccccc}
\toprule
& & \multicolumn{4}{c}{\textbf{ROI 100}} & \multicolumn{4}{c}{\textbf{ROI 400}} & \multicolumn{4}{c}{\textbf{ROI 1000}} & \textbf{Avg. across}  \\
\cmidrule(lr){3-6} \cmidrule(lr){7-10} \cmidrule(lr){11-14}
\textbf{Dataset} &  & \textbf{GCN} & \textbf{GAT} & \textbf{SAGE} & \textbf{Avg. across GNNs} & \textbf{GCN} & \textbf{GAT} & \textbf{SAGE} & \textbf{Avg. across GNNs} & \textbf{GCN} & \textbf{GAT} & \textbf{SAGE} & \textbf{Avg. across GNNs} & \textbf{ROIs \& GNNs} \\
\midrule
\multirow{3}{*}{\rotatebox{90}{ \textbf{Rest~~}}} 
& \textbf{Baseline} & 74.07 ± 1.13 & 74.35 ± 2.56 & 71.11 ± 3.76 & 73.18 & 74.63 ± 0.68 & 76.02 ± 0.80 & 75.37 ± 2.08 & 75.34 & 81.39 ± 2.81 & 78.33 ± 3.43 & 81.48 ± 1.80 & 80.4 & 76.31 \\
& \textbf{Spearman} & 73.06 ± 1.48 & 72.69 ± 2.57 & 73.43 ± 2.69 & 73.06 & 77.69 ± 1.93 & 75.93 ± 2.11 & 76.20 ± 2.30 & 76.61 & 82.13 ± 1.23 & 79.17 ± 5.08 & 80.74 ± 1.19 & 80.68 & 76.78 \\
& \textbf{Kendall} & 74.54 ± 3.10 & 73.70 ± 3.02 & 75.09 ± 1.72 & 74.44 & 78.52 ± 1.42 & 77.13 ± 2.48 & 76.85 ± 2.23 & 77.5 & 84.07 ± 0.86 & 81.02 ± 3.60 & 82.31 ± 1.53 & 82.47 & 78.14 \\
\midrule
\multirow{3}{*}{\rotatebox{90}{ \textbf{Task~~}}} 
& \textbf{Baseline} & 91.05 ± 0.29 & 91.93 ± 0.50 & 91.71 ± 0.30 & 91.56 & 94.57 ± 0.60 & 95.08 ± 0.51 & 95.51 ± 0.19 & 95.05 & 96.06 ± 0.36 & 96.26 ± 0.62 & 96.49 ± 0.28 & 96.27 & 94.3 \\
& \textbf{Spearman} & 92.05 ± 0.51 & 92.09 ± 0.88 & 91.97 ± 0.74 & 92.04 & 94.63 ± 0.09 & 95.10 ± 0.19 & 95.63 ± 0.34 & 95.12 & 96.12 ± 0.49 & 96.84 ± 0.33 & 97.11 ± 0.36 & 96.69 & 94.62 \\
& \textbf{Kendall} & 91.63 ± 0.64 & 92.13 ± 0.70 & 91.97 ± 0.73 & 91.91 & 94.69 ± 0.51 & 95.30 ± 0.36 & 95.61 ± 0.44 & 95.2 & 96.28 ± 0.28 & 96.84 ± 0.22 & 97.18 ± 0.30 & 96.77 & 94.63 \\
\bottomrule
\end{tabular}%
}

\end{table*}

\subsection{Temporal Signal Processing}
As discussed in Section~\ref{sec:denoising}, we treat high-amplitude signal retention as one axis of the data-centric design space for brain graph construction. Here, we examine multiple thresholding strategies—including percentile-based \( \theta = Q_{\alpha}(\mathbf{Z}_r) \) and standard deviation-based \( \theta = \beta \sigma \)—to evaluate whether selectively retaining signal peaks improves the informativeness of functional connectivity patterns.

As shown in Table \ref{tab:denoising_rs}, this particular data-centric design choice, retaining only the top 30\% of signal amplitudes, leads to superior GNN performance compared to using the full BOLD signal. Another strategy we investigated involved setting the threshold at one standard deviation (\(\theta = \sigma\)), applying the function \( \Theta_{\theta_{sd1}\gamma_0} \) to the z-normalized BOLD signal. Additionally, we experimented where the thresholded signal was transformed using \( \Theta_{\theta, \gamma=1} \), mapping values above \( \theta \) to 1 and otherwise 0. In this binary representation, 0 corresponds to sub-threshold signal components, while 1 represents high-amplitude peaks, further refining the feature selection process. Additional observations are as follows:
\vspace{-1.5ex}
\begin{itemize}[left=0pt]
\item In HCP-Rest, the data-centric choice of high-amplitude signal retention notably improves performance, especially at higher ROI resolutions (400 and 1000), where \( \Theta_{\theta, \gamma=1} \) yields substantial gains. Performance benefits align with prior work suggesting high-resolution brain graphs capture finer-grained patterns~\cite{said2023neurograph}. Our signal retention further boosts this by focusing on high-magnitude fluctuations.

\item In HCP-Task, minimal improvement is observed, likely due to already strong task-driven co-activation patterns~\cite{van2013wu}, suggesting limited room for further signal-level optimization.

\item On ABIDE (Table~\ref{tab:denoising_abide}), applying \( \Theta_{\theta_{sd1}\gamma_1} \) improves overall GNN accuracy at ROI 200, with GCN rising from 60.00\% to 65.86\%, demonstrating the effectiveness in clinical datasets.

\end{itemize}

\begin{table}[t]
\centering
\scriptsize
\caption{Classification results on ABIDE for various high-amplitude signal retention strategies.}
\vskip -2ex
\label{tab:denoising_abide}
\begin{tabular}{lccccccc}
\toprule
 & \multicolumn{3}{c}{\textbf{ROI 200}} & \multicolumn{3}{c}{\textbf{ROI 400}} & \textbf{Avg. across}  \\
\cmidrule(lr){2-4} \cmidrule(lr){5-7}
& \textbf{GCN} & \textbf{GAT} & \textbf{SAGE} & \textbf{GCN} & \textbf{GAT} & \textbf{SAGE}  & \textbf{ROIs \& GNNs}\\
\midrule
\textbf{Baseline} & 60.00 & 62.13 & 62.32 &  62.32  & 64.06  & 63.86  & 62.45 \\
{\tiny \textbf{\(\Theta_{\theta_{p30}\gamma_0}\)}} & 61.74  & 62.13  & 62.22  &  61.64  & 63.09  & 57.87  & 61.45 \\
{\tiny\textbf{\(\Theta_{\theta_{p30}\gamma_1}\)}} & 58.94  & 61.45  & 61.35  & 62.13  & 63.96  & 60.00  & 61.31 \\
{\tiny\textbf{\(\Theta_{\theta_{sd1}\gamma_0}\)}} & 62.02  & 62.32  & 61.72  & 60.77  & 62.73  & 61.53  & 61.85 \\
{\tiny\textbf{\(\Theta_{\theta_{sd1}\gamma_1}\)}} & 65.86  & 63.03  & 63.84 & 61.64 & 61.75  & 62.08  & 63.03 \\
\bottomrule
\end{tabular}%
\vskip -2ex
\end{table}

\vspace{-1.5ex}
\subsection{Topology Extraction}
In Section \ref{sec:spearman_kendall}, we explore the use of alternative correlation measures as data-centric design choices for brain graph topology construction. These rank-based metrics aim to better capture non-linear dependencies and reduce sensitivity to outliers compared to Pearson correlation.
As shown in Table~\ref{tab:corr_merged}, graphs constructed with Spearman or Kendall correlations consistently outperform the baseline. We summarizes the findings here as: 
\vspace{-1ex}

\begin{table}[t]
\centering
\scriptsize
\caption{Accuracy of GNNs on ABIDE dataset under different correlation strategies and ROI settings.}
\vskip -2ex
\label{tab:corr_abide}
\begin{tabular}{lccccccc}
\toprule
 & \multicolumn{3}{c}{\textbf{ROI 200}} & \multicolumn{3}{c}{\textbf{ROI 400}} & \textbf{Avg. across} \\
\cmidrule(lr){2-4} \cmidrule(lr){5-7}
& \textbf{GCN} & \textbf{GAT} & \textbf{SAGE} & \textbf{GCN} & \textbf{GAT} & \textbf{SAGE} & \textbf{ ROIs \& GNNs}\\
\midrule
\textbf{Baseline} & 60.00 & 62.13 & 62.32 & 62.32 & 64.06 & 63.86 & 62.45 \\
\textbf{Spearman} & 61.93 & 65.22 & 60.87 & 63.00 & 63.29 & 63.48 & 62.97 \\
\textbf{Kendall} & 62.13 & 63.09 & 61.06 & 63.77 & 63.77 & 63.38 & 62.87 \\
\bottomrule
\end{tabular}
\vskip -2ex
\end{table}

\begin{figure*}[t]
    \centering
    \begin{subfigure}[b]{0.32\textwidth}
        \includegraphics[width=\textwidth]{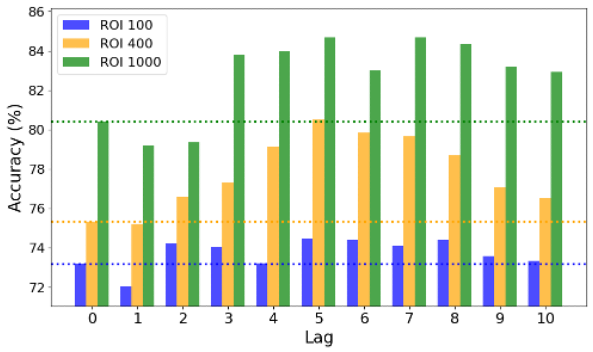}
        \vskip -0.75ex
        \caption{HCP-Rest}
        \label{fig:lag_rest}
    \end{subfigure}
    \begin{subfigure}[b]{0.32\textwidth}
        \includegraphics[width=\textwidth]{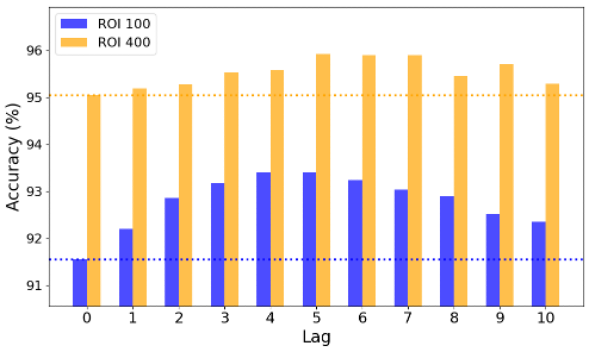}
        \vskip -0.75ex
        \caption{HCP-Task}
        \label{fig:lag_state}
    \end{subfigure}
    \begin{subfigure}[b]{0.32\textwidth}
        \includegraphics[width=\textwidth]{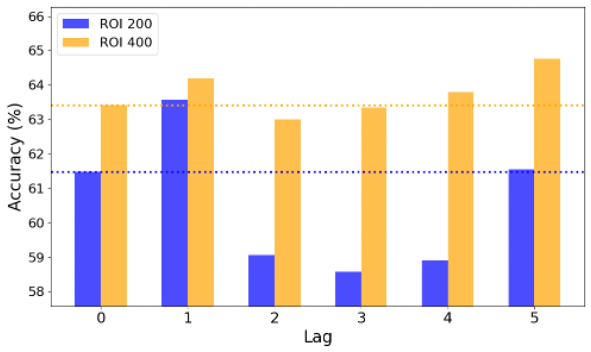}
        \vskip -0.75ex
        \caption{ABIDE}
        \label{fig:lag_ABIDE}
    \end{subfigure}
    \vskip -1.75ex
    \caption{Performance comparison using varying lag/lead correlations across the 
    datasets HCP-Rest, HCP-Task, and ABIDE. }
    \vskip 1ex
    \label{fig:lag}
    \vskip -2.5ex
\end{figure*}

\begin{itemize}[left=0pt]
\item Across all three datasets, using alternative correlation coefficients yields consistent improvements over Pearson, demonstrating the value of selecting appropriate correlation metrics as a data-centric design choice.

\item In HCP-Rest, Kendall correlation achieves the strongest performance, suggesting its effectiveness in capturing informative patterns for functional connectivity. Spearman also consistently outperforms Pearson. Similar trends are observed in HCP-Task (Table~\ref{tab:corr_merged}) and ABIDE (Table~\ref{tab:corr_abide}), reinforcing the utility of rank-based correlations in brain graph construction.

\end{itemize}

\begin{figure}[t]
    \centering
    \includegraphics[width=0.9\linewidth]{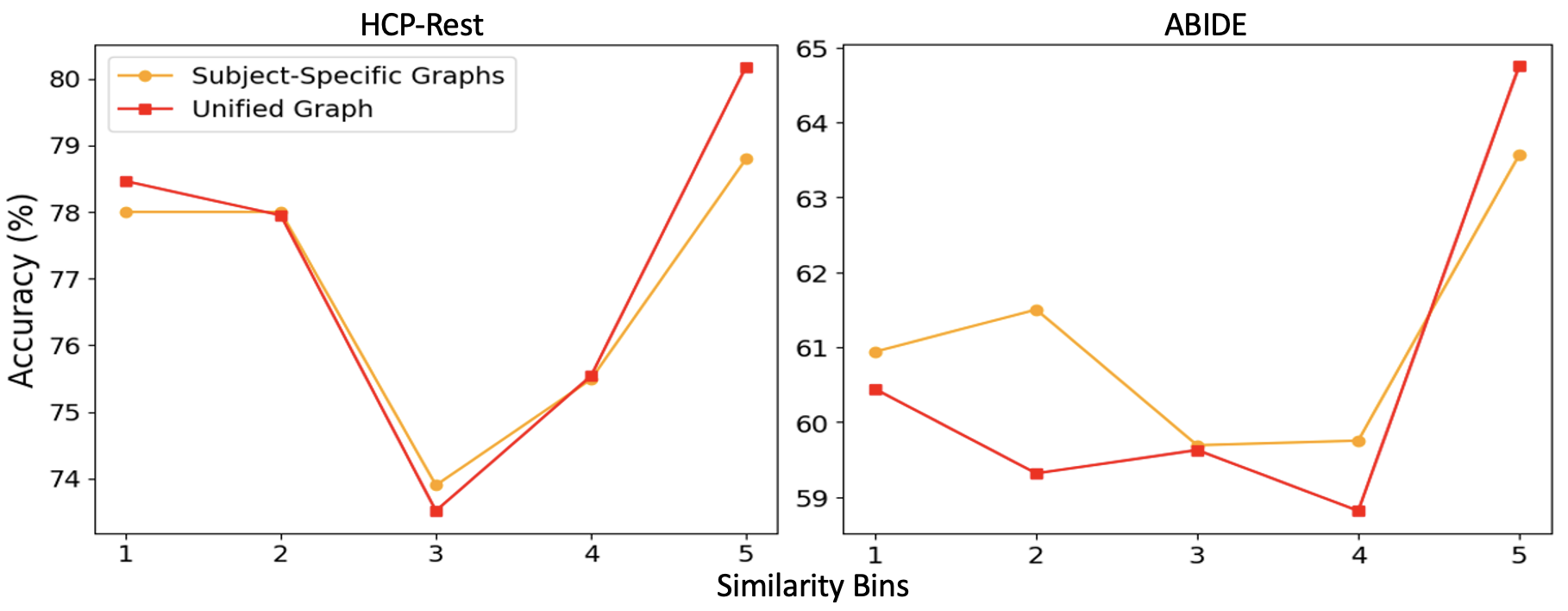}
    \vskip -1.5ex
    \caption{Comparison between subject-specific and unified brain graph topologies on HCP-Rest and ABIDE.}
    \label{fig:unified_graphs}
    \vspace{-2ex}
\end{figure}

In addition, we explore a unified graph topology as a data-centric design choice aimed at enhancing the consistency of brain network representations across subjects. 
We construct global unified graphs as discussed in Section~\ref{sec:unified_graph} and then to evaluate its impact, we stratify test samples into bins based on the overlap between their subject-specific and unified graphs, using edge similarity scores. Due to data imbalance, the lower-similarity bins are combined for analysis. Since task-based fMRI reflects condition-specific co-activation patterns~\cite{van2013wu}, we apply the unified graph only to resting-state datasets. Empirical results, shown in Figure~\ref{fig:unified_graphs}, illustrate the effects of this design choice on downstream performance.
We observe that in the HCP-Rest and ABIDE datasets, the unified topology yields performance gains particularly for samples with high structural similarity to the aggregated graph. These findings highlight that even simple data-centric strategies for enforcing structural consistency—such as shared topology aggregation—can meaningfully enhance GNN performance.

\subsection{Advanced Graph Featurization}

As discussed in Section~\ref{sec:lag}, we explore lagged correlation as a data-centric design choice to incorporate dynamic properties into brain graphs. Rather than relying solely on instantaneous co-activation between ROIs, we augment the node feature matrix by concatenating leading and lagging correlations, capturing temporal dependencies in functional connectivity. As shown in Figure~\ref{fig:lag}, this strategy consistently improves classification performance across datasets. Additional details are provided in Appendix Tables~\ref{tab:8} and \ref{tab:10}, with further analysis of lag effects discussed below.

\vspace{-1ex}
\begin{itemize}[left=0pt]
\item Incorporating lagged correlations consistently improves GNN performance across datasets and settings, suggesting that temporal offsets capture dynamic neural interactions beyond what instantaneous correlation provides.

\item In HCP-Rest and HCP-Task, lag values between 3 and 8 time points provide the most benefit across mental states, especially at higher ROI resolutions where finer-grained dynamics are better captured.

\item In ABIDE, a lag of 1 yields the most improvement, likely due to its longer repetition time (TR $\approx$ 2s)~\cite{di2014autism} compared to HCP (TR $\approx$ 0.72s)~\cite{van2013wu}. This correspondence reinforces the interpretability of lagged correlation as a biologically grounded, data-centric design parameter.

\end{itemize}
\vspace{-1ex}

Additionally, for edge features, we encode multi-perspective connectivity by assigning each edge a binary vector indicating its presence in the Pearson, Spearman, or Kendall correlation graphs, resulting in an edge feature matrix of shape \( |E| \times 3 \). Incorporating these edge-level signals into the standard GNN pipeline yields performance improvements in several settings, as shown in Tables~\ref{tab:edge_feature} and~\ref{tab:edge_feature-abide} in the Appendix~\ref{sec:appendix_table}. To further expand the data-centric design space, we augment node features by concatenating correlation matrices computed from multiple correlation types (Pearson, Spearman, Kendall) and multiple temporal offsets (lag/lead), yielding an enriched node feature matrix of shape \( n \times 5n \). These strategies systematically introduce diverse functional connectivity cues into the graph representation. As detailed in Appendix~\ref{sec:appendix_table}, we observe that when combined with lagged correlation, the most beneficial lag window for HCP lies between 3 and 8 time points, reinforcing its value as a meaningful data-centric hyperparameter.

\section{Conclusion}

In this work, we introduced a data-centric AI framework that systematically explores the design space of brain graph construction from fMRI data. By disentangling and evaluating the effects of feature selection, topology extraction, and graph featurization, we demonstrate consistent improvements in downstream GNN classification across diverse datasets and model settings. Rather than proposing a fixed method, our findings underscore the critical role of data processing choices in shaping brain graph quality and highlight the importance of incorporating these design decisions into the development of effective neuroimaging pipelines.



\balance
\bibliographystyle{ACM-Reference-Format}
\bibliography{references}


\begin{thebibliography}{46}


\ifx \showCODEN    \undefined \def \showCODEN     #1{\unskip}     \fi
\ifx \showISBNx    \undefined \def \showISBNx     #1{\unskip}     \fi
\ifx \showISBNxiii \undefined \def \showISBNxiii  #1{\unskip}     \fi
\ifx \showISSN     \undefined \def \showISSN      #1{\unskip}     \fi
\ifx \showLCCN     \undefined \def \showLCCN      #1{\unskip}     \fi
\ifx \shownote     \undefined \def \shownote      #1{#1}          \fi
\ifx \showarticletitle \undefined \def \showarticletitle #1{#1}   \fi
\ifx \showURL      \undefined \def \showURL       {\relax}        \fi
\providecommand\bibfield[2]{#2}
\providecommand\bibinfo[2]{#2}
\providecommand\natexlab[1]{#1}
\providecommand\showeprint[2][]{arXiv:#2}

\bibitem[Ahmadi et~al\mbox{.}(2023)]%
        {ahmadi2023comparative}
\bibfield{author}{\bibinfo{person}{Hessam Ahmadi}, \bibinfo{person}{Emad Fatemizadeh}, {and} \bibinfo{person}{Ali Motie-Nasrabadi}.} \bibinfo{year}{2023}\natexlab{}.
\newblock \showarticletitle{A comparative study of correlation methods in functional connectivity analysis using fMRI data of alzheimer’s patients}.
\newblock \bibinfo{journal}{\emph{Journal of Biomedical Physics \& Engineering}} \bibinfo{volume}{13}, \bibinfo{number}{2} (\bibinfo{year}{2023}), \bibinfo{pages}{125}.
\newblock


\bibitem[Arbabshirani et~al\mbox{.}(2017)]%
        {arbabshirani2017single}
\bibfield{author}{\bibinfo{person}{Mohammad~R Arbabshirani}, \bibinfo{person}{Sergey Plis}, \bibinfo{person}{Jing Sui}, {and} \bibinfo{person}{Vince~D Calhoun}.} \bibinfo{year}{2017}\natexlab{}.
\newblock \showarticletitle{Single subject prediction of brain disorders in neuroimaging: Promises and pitfalls}.
\newblock \bibinfo{journal}{\emph{Neuroimage}}  \bibinfo{volume}{145} (\bibinfo{year}{2017}), \bibinfo{pages}{137--165}.
\newblock


\bibitem[Caballero-Gaudes and Reynolds(2017)]%
        {caballero2017methods}
\bibfield{author}{\bibinfo{person}{C{\'e}sar Caballero-Gaudes} {and} \bibinfo{person}{Richard~C Reynolds}.} \bibinfo{year}{2017}\natexlab{}.
\newblock \showarticletitle{Methods for cleaning the BOLD fMRI signal}.
\newblock \bibinfo{journal}{\emph{Neuroimage}}  \bibinfo{volume}{154} (\bibinfo{year}{2017}), \bibinfo{pages}{128--149}.
\newblock


\bibitem[Craddock et~al\mbox{.}(2012)]%
        {craddock2012whole}
\bibfield{author}{\bibinfo{person}{R~Cameron Craddock}, \bibinfo{person}{G~Andrew James}, \bibinfo{person}{Paul~E Holtzheimer~III}, \bibinfo{person}{Xiaoping~P Hu}, {and} \bibinfo{person}{Helen~S Mayberg}.} \bibinfo{year}{2012}\natexlab{}.
\newblock \showarticletitle{A whole brain fMRI atlas generated via spatially constrained spectral clustering}.
\newblock \bibinfo{journal}{\emph{Human brain mapping}} \bibinfo{volume}{33}, \bibinfo{number}{8} (\bibinfo{year}{2012}), \bibinfo{pages}{1914--1928}.
\newblock


\bibitem[Cui et~al\mbox{.}(2022a)]%
        {cui2022braingb}
\bibfield{author}{\bibinfo{person}{Hejie Cui}, \bibinfo{person}{Wei Dai}, \bibinfo{person}{Yanqiao Zhu}, \bibinfo{person}{Xuan Kan}, \bibinfo{person}{Antonio Aodong~Chen Gu}, \bibinfo{person}{Joshua Lukemire}, \bibinfo{person}{Liang Zhan}, \bibinfo{person}{Lifang He}, \bibinfo{person}{Ying Guo}, {and} \bibinfo{person}{Carl Yang}.} \bibinfo{year}{2022}\natexlab{a}.
\newblock \showarticletitle{Braingb: a benchmark for brain network analysis with graph neural networks}.
\newblock \bibinfo{journal}{\emph{IEEE transactions on medical imaging}} \bibinfo{volume}{42}, \bibinfo{number}{2} (\bibinfo{year}{2022}), \bibinfo{pages}{493--506}.
\newblock


\bibitem[Cui et~al\mbox{.}(2022b)]%
        {cui2022interpretable}
\bibfield{author}{\bibinfo{person}{Hejie Cui}, \bibinfo{person}{Wei Dai}, \bibinfo{person}{Yanqiao Zhu}, \bibinfo{person}{Xiaoxiao Li}, \bibinfo{person}{Lifang He}, {and} \bibinfo{person}{Carl Yang}.} \bibinfo{year}{2022}\natexlab{b}.
\newblock \showarticletitle{Interpretable graph neural networks for connectome-based brain disorder analysis}. In \bibinfo{booktitle}{\emph{International conference on medical image computing and computer-assisted intervention}}. Springer, \bibinfo{pages}{375--385}.
\newblock


\bibitem[Dahan et~al\mbox{.}(2021)]%
        {dahan2021improving}
\bibfield{author}{\bibinfo{person}{Simon Dahan}, \bibinfo{person}{Logan~ZJ Williams}, \bibinfo{person}{Daniel Rueckert}, {and} \bibinfo{person}{Emma~C Robinson}.} \bibinfo{year}{2021}\natexlab{}.
\newblock \showarticletitle{Improving phenotype prediction using long-range spatio-temporal dynamics of functional connectivity}. In \bibinfo{booktitle}{\emph{Machine Learning in Clinical Neuroimaging: 4th International Workshop, MLCN 2021, Held in Conjunction with MICCAI 2021, Strasbourg, France, September 27, 2021, Proceedings 4}}. Springer, \bibinfo{pages}{145--154}.
\newblock


\bibitem[Di~Martino et~al\mbox{.}(2014)]%
        {di2014autism}
\bibfield{author}{\bibinfo{person}{Adriana Di~Martino}, \bibinfo{person}{Chao-Gan Yan}, \bibinfo{person}{Qingyang Li}, \bibinfo{person}{Erin Denio}, \bibinfo{person}{Francisco~X Castellanos}, \bibinfo{person}{Kaat Alaerts}, \bibinfo{person}{Jeffrey~S Anderson}, \bibinfo{person}{Michal Assaf}, \bibinfo{person}{Susan~Y Bookheimer}, \bibinfo{person}{Mirella Dapretto}, {et~al\mbox{.}}} \bibinfo{year}{2014}\natexlab{}.
\newblock \showarticletitle{The autism brain imaging data exchange: towards a large-scale evaluation of the intrinsic brain architecture in autism}.
\newblock \bibinfo{journal}{\emph{Molecular psychiatry}} \bibinfo{volume}{19}, \bibinfo{number}{6} (\bibinfo{year}{2014}), \bibinfo{pages}{659--667}.
\newblock


\bibitem[Du et~al\mbox{.}(2024)]%
        {du2024survey}
\bibfield{author}{\bibinfo{person}{Yuhui Du}, \bibinfo{person}{Songke Fang}, \bibinfo{person}{Xingyu He}, {and} \bibinfo{person}{Vince~D Calhoun}.} \bibinfo{year}{2024}\natexlab{}.
\newblock \showarticletitle{A survey of brain functional network extraction methods using fMRI data}.
\newblock \bibinfo{journal}{\emph{Trends in Neurosciences}} (\bibinfo{year}{2024}).
\newblock


\bibitem[Fiecas et~al\mbox{.}(2013)]%
        {fiecas2013quantifying}
\bibfield{author}{\bibinfo{person}{Mark Fiecas}, \bibinfo{person}{Hernando Ombao}, \bibinfo{person}{Dan Van~Lunen}, \bibinfo{person}{Richard Baumgartner}, \bibinfo{person}{Alexandre Coimbra}, {and} \bibinfo{person}{Dai Feng}.} \bibinfo{year}{2013}\natexlab{}.
\newblock \showarticletitle{Quantifying temporal correlations: A test--retest evaluation of functional connectivity in resting-state fMRI}.
\newblock \bibinfo{journal}{\emph{NeuroImage}}  \bibinfo{volume}{65} (\bibinfo{year}{2013}).
\newblock


\bibitem[Gadgil et~al\mbox{.}(2020)]%
        {gadgil2020spatio}
\bibfield{author}{\bibinfo{person}{Soham Gadgil}, \bibinfo{person}{Qingyu Zhao}, \bibinfo{person}{Adolf Pfefferbaum}, \bibinfo{person}{Edith~V Sullivan}, \bibinfo{person}{Ehsan Adeli}, {and} \bibinfo{person}{Kilian~M Pohl}.} \bibinfo{year}{2020}\natexlab{}.
\newblock \showarticletitle{Spatio-temporal graph convolution for resting-state fMRI analysis}. In \bibinfo{booktitle}{\emph{Medical Image Computing and Computer Assisted Intervention--MICCAI 2020: 23rd International Conference, Lima, Peru, October 4--8, 2020, Proceedings, Part VII 23}}. Springer, \bibinfo{pages}{528--538}.
\newblock


\bibitem[Glasser et~al\mbox{.}(2013)]%
        {glasser2013minimal}
\bibfield{author}{\bibinfo{person}{Matthew~F Glasser}, \bibinfo{person}{Stamatios~N Sotiropoulos}, \bibinfo{person}{J~Anthony Wilson}, \bibinfo{person}{Timothy~S Coalson}, \bibinfo{person}{Bruce Fischl}, \bibinfo{person}{Jesper~L Andersson}, \bibinfo{person}{Junqian Xu}, \bibinfo{person}{Saad Jbabdi}, \bibinfo{person}{Matthew Webster}, \bibinfo{person}{Jonathan~R Polimeni}, {et~al\mbox{.}}} \bibinfo{year}{2013}\natexlab{}.
\newblock \showarticletitle{The minimal preprocessing pipelines for the Human Connectome Project}.
\newblock \bibinfo{journal}{\emph{Neuroimage}}  \bibinfo{volume}{80} (\bibinfo{year}{2013}), \bibinfo{pages}{105--124}.
\newblock


\bibitem[Hyde and Jesmanowicz(2012)]%
        {hyde2012cross}
\bibfield{author}{\bibinfo{person}{James~S Hyde} {and} \bibinfo{person}{Andrzej Jesmanowicz}.} \bibinfo{year}{2012}\natexlab{}.
\newblock \showarticletitle{Cross-correlation: an fMRI signal-processing strategy}.
\newblock \bibinfo{journal}{\emph{NeuroImage}} \bibinfo{volume}{62}, \bibinfo{number}{2} (\bibinfo{year}{2012}), \bibinfo{pages}{848--851}.
\newblock


\bibitem[Jakubik et~al\mbox{.}(2024)]%
        {jakubik2024data}
\bibfield{author}{\bibinfo{person}{Johannes Jakubik}, \bibinfo{person}{Michael V{\"o}ssing}, \bibinfo{person}{Niklas K{\"u}hl}, \bibinfo{person}{Jannis Walk}, {and} \bibinfo{person}{Gerhard Satzger}.} \bibinfo{year}{2024}\natexlab{}.
\newblock \showarticletitle{Data-centric artificial intelligence}.
\newblock \bibinfo{journal}{\emph{Business \& Information Systems Engineering}} (\bibinfo{year}{2024}), \bibinfo{pages}{1--9}.
\newblock


\bibitem[Jarrahi et~al\mbox{.}(2022)]%
        {jarrahi2022principles}
\bibfield{author}{\bibinfo{person}{Mohammad~Hossein Jarrahi}, \bibinfo{person}{Ali Memariani}, {and} \bibinfo{person}{Shion Guha}.} \bibinfo{year}{2022}\natexlab{}.
\newblock \showarticletitle{The principles of data-centric AI (DCAI)}.
\newblock \bibinfo{journal}{\emph{arXiv preprint arXiv:2211.14611}} (\bibinfo{year}{2022}).
\newblock


\bibitem[Kim et~al\mbox{.}(2021)]%
        {kim2021learning}
\bibfield{author}{\bibinfo{person}{Byung-Hoon Kim}, \bibinfo{person}{Jong~Chul Ye}, {and} \bibinfo{person}{Jae-Jin Kim}.} \bibinfo{year}{2021}\natexlab{}.
\newblock \showarticletitle{Learning dynamic graph representation of brain connectome with spatio-temporal attention}.
\newblock \bibinfo{journal}{\emph{Advances in Neural Information Processing Systems}}  \bibinfo{volume}{34} (\bibinfo{year}{2021}), \bibinfo{pages}{4314--4327}.
\newblock


\bibitem[LeCun et~al\mbox{.}(2015)]%
        {lecun2015deep}
\bibfield{author}{\bibinfo{person}{Yann LeCun}, \bibinfo{person}{Yoshua Bengio}, {and} \bibinfo{person}{Geoffrey Hinton}.} \bibinfo{year}{2015}\natexlab{}.
\newblock \showarticletitle{Deep learning}.
\newblock \bibinfo{journal}{\emph{nature}} \bibinfo{volume}{521}, \bibinfo{number}{7553} (\bibinfo{year}{2015}), \bibinfo{pages}{436--444}.
\newblock


\bibitem[Li et~al\mbox{.}(2023)]%
        {li2023interpretable}
\bibfield{author}{\bibinfo{person}{Gaotang Li}, \bibinfo{person}{Marlena Duda}, \bibinfo{person}{Xiang Zhang}, \bibinfo{person}{Danai Koutra}, {and} \bibinfo{person}{Yujun Yan}.} \bibinfo{year}{2023}\natexlab{}.
\newblock \showarticletitle{Interpretable sparsification of brain graphs: Better practices and effective designs for graph neural networks}. In \bibinfo{booktitle}{\emph{Proceedings of the 29th ACM SIGKDD Conference on Knowledge Discovery and Data Mining}}. \bibinfo{pages}{1223--1234}.
\newblock


\bibitem[Li et~al\mbox{.}(2021)]%
        {li2021braingnn}
\bibfield{author}{\bibinfo{person}{Xiaoxiao Li}, \bibinfo{person}{Yuan Zhou}, \bibinfo{person}{Nicha Dvornek}, \bibinfo{person}{Muhan Zhang}, \bibinfo{person}{Siyuan Gao}, \bibinfo{person}{Juntang Zhuang}, \bibinfo{person}{Dustin Scheinost}, \bibinfo{person}{Lawrence~H Staib}, \bibinfo{person}{Pamela Ventola}, {and} \bibinfo{person}{James~S Duncan}.} \bibinfo{year}{2021}\natexlab{}.
\newblock \showarticletitle{Braingnn: Interpretable brain graph neural network for fmri analysis}.
\newblock \bibinfo{journal}{\emph{Medical Image Analysis}}  \bibinfo{volume}{74} (\bibinfo{year}{2021}), \bibinfo{pages}{102233}.
\newblock


\bibitem[Liu and Duyn(2013)]%
        {liu2013time}
\bibfield{author}{\bibinfo{person}{Xiao Liu} {and} \bibinfo{person}{Jeff~H Duyn}.} \bibinfo{year}{2013}\natexlab{}.
\newblock \showarticletitle{Time-varying functional network information extracted from brief instances of spontaneous brain activity}.
\newblock \bibinfo{journal}{\emph{Proceedings of the National Academy of Sciences}} \bibinfo{volume}{110}, \bibinfo{number}{11} (\bibinfo{year}{2013}), \bibinfo{pages}{4392--4397}.
\newblock


\bibitem[Luo et~al\mbox{.}(2024)]%
        {luo2024graph}
\bibfield{author}{\bibinfo{person}{Xuexiong Luo}, \bibinfo{person}{Jia Wu}, \bibinfo{person}{Jian Yang}, \bibinfo{person}{Shan Xue}, \bibinfo{person}{Amin Beheshti}, \bibinfo{person}{Quan~Z Sheng}, \bibinfo{person}{David McAlpine}, \bibinfo{person}{Paul Sowman}, \bibinfo{person}{Alexis Giral}, {and} \bibinfo{person}{Philip~S Yu}.} \bibinfo{year}{2024}\natexlab{}.
\newblock \showarticletitle{Graph Neural Networks for Brain Graph Learning: A Survey}.
\newblock \bibinfo{journal}{\emph{arXiv preprint arXiv:2406.02594}} (\bibinfo{year}{2024}).
\newblock


\bibitem[Marrelec et~al\mbox{.}(2006)]%
        {marrelec2006partial}
\bibfield{author}{\bibinfo{person}{Guillaume Marrelec}, \bibinfo{person}{Alexandre Krainik}, \bibinfo{person}{Hugues Duffau}, \bibinfo{person}{M{\'e}lanie P{\'e}l{\'e}grini-Issac}, \bibinfo{person}{St{\'e}phane Leh{\'e}ricy}, \bibinfo{person}{Julien Doyon}, {and} \bibinfo{person}{Habib Benali}.} \bibinfo{year}{2006}\natexlab{}.
\newblock \showarticletitle{Partial correlation for functional brain interactivity investigation in functional MRI}.
\newblock \bibinfo{journal}{\emph{Neuroimage}} \bibinfo{volume}{32}, \bibinfo{number}{1} (\bibinfo{year}{2006}), \bibinfo{pages}{228--237}.
\newblock


\bibitem[Ng(2023)]%
        {ng2025LandingAI}
\bibfield{author}{\bibinfo{person}{Andrew Ng}.} \bibinfo{year}{2023}\natexlab{}.
\newblock \bibinfo{title}{Landing AI}.
\newblock \bibinfo{howpublished}{\url{https://landing.ai/}}.
\newblock
\newblock
\shownote{Retrieved February 8, 2025}.


\bibitem[Plis et~al\mbox{.}(2014)]%
        {plis2014deep}
\bibfield{author}{\bibinfo{person}{Sergey~M Plis}, \bibinfo{person}{Devon~R Hjelm}, \bibinfo{person}{Ruslan Salakhutdinov}, \bibinfo{person}{Elena~A Allen}, \bibinfo{person}{Henry~J Bockholt}, \bibinfo{person}{Jeffrey~D Long}, \bibinfo{person}{Hans~J Johnson}, \bibinfo{person}{Jane~S Paulsen}, \bibinfo{person}{Jessica~A Turner}, {and} \bibinfo{person}{Vince~D Calhoun}.} \bibinfo{year}{2014}\natexlab{}.
\newblock \showarticletitle{Deep learning for neuroimaging: a validation study}.
\newblock \bibinfo{journal}{\emph{Frontiers in neuroscience}}  \bibinfo{volume}{8} (\bibinfo{year}{2014}), \bibinfo{pages}{229}.
\newblock


\bibitem[Power et~al\mbox{.}(2017)]%
        {power2017sources}
\bibfield{author}{\bibinfo{person}{Jonathan~D Power}, \bibinfo{person}{Mark Plitt}, \bibinfo{person}{Timothy~O Laumann}, {and} \bibinfo{person}{Alex Martin}.} \bibinfo{year}{2017}\natexlab{}.
\newblock \showarticletitle{Sources and implications of whole-brain fMRI signals in humans}.
\newblock \bibinfo{journal}{\emph{Neuroimage}}  \bibinfo{volume}{146} (\bibinfo{year}{2017}), \bibinfo{pages}{609--625}.
\newblock


\bibitem[Rousselet and Pernet(2012)]%
        {rousselet2012improving}
\bibfield{author}{\bibinfo{person}{Guillaume~A Rousselet} {and} \bibinfo{person}{Cyril~R Pernet}.} \bibinfo{year}{2012}\natexlab{}.
\newblock \showarticletitle{Improving standards in brain-behavior correlation analyses}.
\newblock \bibinfo{journal}{\emph{Frontiers in human neuroscience}}  \bibinfo{volume}{6} (\bibinfo{year}{2012}), \bibinfo{pages}{119}.
\newblock


\bibitem[Said et~al\mbox{.}(2023)]%
        {said2023neurograph}
\bibfield{author}{\bibinfo{person}{Anwar Said}, \bibinfo{person}{Roza Bayrak}, \bibinfo{person}{Tyler Derr}, \bibinfo{person}{Mudassir Shabbir}, \bibinfo{person}{Daniel Moyer}, \bibinfo{person}{Catie Chang}, {and} \bibinfo{person}{Xenofon Koutsoukos}.} \bibinfo{year}{2023}\natexlab{}.
\newblock \showarticletitle{Neurograph: Benchmarks for graph machine learning in brain connectomics}.
\newblock \bibinfo{journal}{\emph{Advances in Neural Information Processing Systems}}  \bibinfo{volume}{36} (\bibinfo{year}{2023}), \bibinfo{pages}{6509--6531}.
\newblock


\bibitem[Schaefer et~al\mbox{.}(2018)]%
        {schaefer2018local}
\bibfield{author}{\bibinfo{person}{Alexander Schaefer}, \bibinfo{person}{Ru Kong}, \bibinfo{person}{Evan~M Gordon}, \bibinfo{person}{Timothy~O Laumann}, \bibinfo{person}{Xi-Nian Zuo}, \bibinfo{person}{Avram~J Holmes}, \bibinfo{person}{Simon~B Eickhoff}, {and} \bibinfo{person}{BT~Thomas Yeo}.} \bibinfo{year}{2018}\natexlab{}.
\newblock \showarticletitle{Local-global parcellation of the human cerebral cortex from intrinsic functional connectivity MRI}.
\newblock \bibinfo{journal}{\emph{Cerebral cortex}} \bibinfo{volume}{28}, \bibinfo{number}{9} (\bibinfo{year}{2018}), \bibinfo{pages}{3095--3114}.
\newblock


\bibitem[Singh(2023)]%
        {singh2023systematic}
\bibfield{author}{\bibinfo{person}{Prerna Singh}.} \bibinfo{year}{2023}\natexlab{}.
\newblock \showarticletitle{Systematic review of data-centric approaches in artificial intelligence and machine learning}.
\newblock \bibinfo{journal}{\emph{Data Science and Management}} \bibinfo{volume}{6}, \bibinfo{number}{3} (\bibinfo{year}{2023}).
\newblock


\bibitem[Sivgin et~al\mbox{.}(2024)]%
        {sivgin2024plug}
\bibfield{author}{\bibinfo{person}{Irmak Sivgin}, \bibinfo{person}{Hasan~Atakan Bedel}, \bibinfo{person}{Saban Ozturk}, {and} \bibinfo{person}{Tolga {\c{C}}ukur}.} \bibinfo{year}{2024}\natexlab{}.
\newblock \showarticletitle{A plug-in graph neural network to boost temporal sensitivity in fmri analysis}.
\newblock \bibinfo{journal}{\emph{IEEE Journal of Biomedical and Health Informatics}} (\bibinfo{year}{2024}).
\newblock


\bibitem[Smith(2004)]%
        {smith2004overview}
\bibfield{author}{\bibinfo{person}{Stephen~M Smith}.} \bibinfo{year}{2004}\natexlab{}.
\newblock \showarticletitle{Overview of fMRI analysis}.
\newblock \bibinfo{journal}{\emph{The British Journal of Radiology}} \bibinfo{volume}{77}, \bibinfo{number}{suppl\_2} (\bibinfo{year}{2004}), \bibinfo{pages}{S167--S175}.
\newblock


\bibitem[Smith et~al\mbox{.}(2011)]%
        {smith2011network}
\bibfield{author}{\bibinfo{person}{Stephen~M Smith}, \bibinfo{person}{Karla~L Miller}, \bibinfo{person}{Gholamreza Salimi-Khorshidi}, \bibinfo{person}{Matthew Webster}, \bibinfo{person}{Christian~F Beckmann}, \bibinfo{person}{Thomas~E Nichols}, \bibinfo{person}{Joseph~D Ramsey}, {and} \bibinfo{person}{Mark~W Woolrich}.} \bibinfo{year}{2011}\natexlab{}.
\newblock \showarticletitle{Network modelling methods for FMRI}.
\newblock \bibinfo{journal}{\emph{Neuroimage}} \bibinfo{volume}{54}, \bibinfo{number}{2} (\bibinfo{year}{2011}).
\newblock


\bibitem[Smith et~al\mbox{.}(2013)]%
        {smith2013functional}
\bibfield{author}{\bibinfo{person}{Stephen~M Smith}, \bibinfo{person}{Diego Vidaurre}, \bibinfo{person}{Christian~F Beckmann}, \bibinfo{person}{Matthew~F Glasser}, \bibinfo{person}{Mark Jenkinson}, \bibinfo{person}{Karla~L Miller}, \bibinfo{person}{Thomas~E Nichols}, \bibinfo{person}{Emma~C Robinson}, \bibinfo{person}{Gholamreza Salimi-Khorshidi}, \bibinfo{person}{Mark~W Woolrich}, {et~al\mbox{.}}} \bibinfo{year}{2013}\natexlab{}.
\newblock \showarticletitle{Functional connectomics from resting-state fMRI}.
\newblock \bibinfo{journal}{\emph{Trends in cognitive sciences}} \bibinfo{volume}{17}, \bibinfo{number}{12} (\bibinfo{year}{2013}), \bibinfo{pages}{666--682}.
\newblock


\bibitem[Tagliazucchi et~al\mbox{.}(2012)]%
        {tagliazucchi2012criticality}
\bibfield{author}{\bibinfo{person}{Enzo Tagliazucchi}, \bibinfo{person}{Pablo Balenzuela}, \bibinfo{person}{Daniel Fraiman}, {and} \bibinfo{person}{Dante~R Chialvo}.} \bibinfo{year}{2012}\natexlab{}.
\newblock \showarticletitle{Criticality in large-scale brain fMRI dynamics unveiled by a novel point process analysis}.
\newblock \bibinfo{journal}{\emph{Frontiers in physiology}}  \bibinfo{volume}{3} (\bibinfo{year}{2012}), \bibinfo{pages}{15}.
\newblock


\bibitem[Van~Essen et~al\mbox{.}(2013)]%
        {van2013wu}
\bibfield{author}{\bibinfo{person}{David~C Van~Essen}, \bibinfo{person}{Stephen~M Smith}, \bibinfo{person}{Deanna~M Barch}, \bibinfo{person}{Timothy~EJ Behrens}, \bibinfo{person}{Essa Yacoub}, \bibinfo{person}{Kamil Ugurbil}, \bibinfo{person}{Wu-Minn~HCP Consortium}, {et~al\mbox{.}}} \bibinfo{year}{2013}\natexlab{}.
\newblock \showarticletitle{The WU-Minn human connectome project: an overview}.
\newblock \bibinfo{journal}{\emph{Neuroimage}}  \bibinfo{volume}{80} (\bibinfo{year}{2013}), \bibinfo{pages}{62--79}.
\newblock


\bibitem[Vieira et~al\mbox{.}(2017)]%
        {vieira2017using}
\bibfield{author}{\bibinfo{person}{Sandra Vieira}, \bibinfo{person}{Walter~HL Pinaya}, {and} \bibinfo{person}{Andrea Mechelli}.} \bibinfo{year}{2017}\natexlab{}.
\newblock \showarticletitle{Using deep learning to investigate the neuroimaging correlates of psychiatric and neurological disorders: Methods and applications}.
\newblock \bibinfo{journal}{\emph{Neuroscience \& Biobehavioral Reviews}}  \bibinfo{volume}{74}.
\newblock


\bibitem[Wang et~al\mbox{.}(2022)]%
        {wang2022improving}
\bibfield{author}{\bibinfo{person}{Yu Wang}, \bibinfo{person}{Yuying Zhao}, \bibinfo{person}{Yushun Dong}, \bibinfo{person}{Huiyuan Chen}, \bibinfo{person}{Jundong Li}, {and} \bibinfo{person}{Tyler Derr}.} \bibinfo{year}{2022}\natexlab{}.
\newblock \showarticletitle{Improving fairness in graph neural networks via mitigating sensitive attribute leakage}. In \bibinfo{booktitle}{\emph{Proceedings of the 28th ACM SIGKDD conference on knowledge discovery and data mining}}. \bibinfo{pages}{1938--1948}.
\newblock


\bibitem[Wilcox and Rousselet(2023)]%
        {wilcox2023updated}
\bibfield{author}{\bibinfo{person}{Rand~R Wilcox} {and} \bibinfo{person}{Guillaume~A Rousselet}.} \bibinfo{year}{2023}\natexlab{}.
\newblock \showarticletitle{An updated guide to robust statistical methods in neuroscience}.
\newblock \bibinfo{journal}{\emph{Current Protocols}} \bibinfo{volume}{3}, \bibinfo{number}{3} (\bibinfo{year}{2023}), \bibinfo{pages}{e719}.
\newblock


\bibitem[Wu et~al\mbox{.}(2020)]%
        {wu2020comprehensive}
\bibfield{author}{\bibinfo{person}{Zonghan Wu}, \bibinfo{person}{Shirui Pan}, \bibinfo{person}{Fengwen Chen}, \bibinfo{person}{Guodong Long}, \bibinfo{person}{Chengqi Zhang}, {and} \bibinfo{person}{S~Yu Philip}.} \bibinfo{year}{2020}\natexlab{}.
\newblock \showarticletitle{A comprehensive survey on graph neural networks}.
\newblock \bibinfo{journal}{\emph{IEEE transactions on neural networks and learning systems}} \bibinfo{volume}{32}, \bibinfo{number}{1} (\bibinfo{year}{2020}), \bibinfo{pages}{4--24}.
\newblock


\bibitem[Yan et~al\mbox{.}(2022)]%
        {yan2022deep}
\bibfield{author}{\bibinfo{person}{Weizheng Yan}, \bibinfo{person}{Gang Qu}, \bibinfo{person}{Wenxing Hu}, \bibinfo{person}{Anees Abrol}, \bibinfo{person}{Biao Cai}, \bibinfo{person}{Chen Qiao}, \bibinfo{person}{Sergey~M Plis}, \bibinfo{person}{Yu-Ping Wang}, \bibinfo{person}{Jing Sui}, {and} \bibinfo{person}{Vince~D Calhoun}.} \bibinfo{year}{2022}\natexlab{}.
\newblock \showarticletitle{Deep learning in neuroimaging: Promises and challenges}.
\newblock \bibinfo{journal}{\emph{IEEE Signal Processing Magazine}} \bibinfo{volume}{39}, \bibinfo{number}{2} (\bibinfo{year}{2022}), \bibinfo{pages}{87--98}.
\newblock


\bibitem[Yang et~al\mbox{.}(2023)]%
        {yang2023data}
\bibfield{author}{\bibinfo{person}{Cheng Yang}, \bibinfo{person}{Deyu Bo}, \bibinfo{person}{Jixi Liu}, \bibinfo{person}{Yufei Peng}, \bibinfo{person}{Boyu Chen}, \bibinfo{person}{Haoran Dai}, \bibinfo{person}{Ao Sun}, \bibinfo{person}{Yue Yu}, \bibinfo{person}{Yixin Xiao}, \bibinfo{person}{Qi Zhang}, {et~al\mbox{.}}} \bibinfo{year}{2023}\natexlab{}.
\newblock \showarticletitle{Data-centric graph learning: A survey}.
\newblock \bibinfo{journal}{\emph{arXiv preprint arXiv:2310.04987}} (\bibinfo{year}{2023}).
\newblock


\bibitem[You et~al\mbox{.}(2020)]%
        {you2020design}
\bibfield{author}{\bibinfo{person}{Jiaxuan You}, \bibinfo{person}{Zhitao Ying}, {and} \bibinfo{person}{Jure Leskovec}.} \bibinfo{year}{2020}\natexlab{}.
\newblock \showarticletitle{Design space for graph neural networks}.
\newblock \bibinfo{journal}{\emph{Advances in Neural Information Processing Systems}}  \bibinfo{volume}{33} (\bibinfo{year}{2020}), \bibinfo{pages}{17009--17021}.
\newblock


\bibitem[Zha et~al\mbox{.}(2023)]%
        {zha2023data}
\bibfield{author}{\bibinfo{person}{Daochen Zha}, \bibinfo{person}{Zaid~Pervaiz Bhat}, \bibinfo{person}{Kwei-Herng Lai}, \bibinfo{person}{Fan Yang}, {and} \bibinfo{person}{Xia Hu}.} \bibinfo{year}{2023}\natexlab{}.
\newblock \showarticletitle{Data-centric ai: Perspectives and challenges}. In \bibinfo{booktitle}{\emph{Proceedings of the 2023 SIAM International Conference on Data Mining (SDM)}}. SIAM, \bibinfo{pages}{945--948}.
\newblock


\bibitem[Zha et~al\mbox{.}(2025)]%
        {zha2025data}
\bibfield{author}{\bibinfo{person}{Daochen Zha}, \bibinfo{person}{Zaid~Pervaiz Bhat}, \bibinfo{person}{Kwei-Herng Lai}, \bibinfo{person}{Fan Yang}, \bibinfo{person}{Zhimeng Jiang}, \bibinfo{person}{Shaochen Zhong}, {and} \bibinfo{person}{Xia Hu}.} \bibinfo{year}{2025}\natexlab{}.
\newblock \showarticletitle{Data-centric artificial intelligence: A survey}.
\newblock \bibinfo{journal}{\emph{Comput. Surveys}} \bibinfo{volume}{57}, \bibinfo{number}{5} (\bibinfo{year}{2025}), \bibinfo{pages}{1--42}.
\newblock


\bibitem[Zhang et~al\mbox{.}(2018)]%
        {zhang2018end}
\bibfield{author}{\bibinfo{person}{Muhan Zhang}, \bibinfo{person}{Zhicheng Cui}, \bibinfo{person}{Marion Neumann}, {and} \bibinfo{person}{Yixin Chen}.} \bibinfo{year}{2018}\natexlab{}.
\newblock \showarticletitle{An end-to-end deep learning architecture for graph classification}. In \bibinfo{booktitle}{\emph{Proceedings of the AAAI conference on artificial intelligence}}, Vol.~\bibinfo{volume}{32}.
\newblock


\bibitem[Zhu et~al\mbox{.}(2024)]%
        {zhu2024spatio}
\bibfield{author}{\bibinfo{person}{Qi Zhu}, \bibinfo{person}{Shengrong Li}, \bibinfo{person}{Xiangshui Meng}, \bibinfo{person}{Qiang Xu}, \bibinfo{person}{Zhiqiang Zhang}, \bibinfo{person}{Wei Shao}, {and} \bibinfo{person}{Daoqiang Zhang}.} \bibinfo{year}{2024}\natexlab{}.
\newblock \showarticletitle{Spatio-Temporal Graph Hubness Propagation Model for Dynamic Brain Network Classification}.
\newblock \bibinfo{journal}{\emph{IEEE Transactions on Medical Imaging}}.
\newblock


\end{thebibliography}

\clearpage
\twocolumn
\appendix
\section{Notations}\label{sec:notations}
Here, in Table~\ref{tab:notations}, we provide a concise listing of the primary notations used in this work along with a short description. 

\begin{table}[h]
    \centering
    \footnotesize 
    \caption{Notations and corresponding descriptions. }
    \vspace{-2ex}
    \label{tab:notations}
    \renewcommand{\arraystretch}{1.3}
    \begin{tabular}{l l}
        \hline
        \textbf{Notations} & \textbf{Description} \\ 
        \hline
        \( N \) & Number of samples \\
        \( T \) & Number of time points in the fMRI signal \\
        $v_i \in \mathcal{V}$ ($r_i \in \mathcal{R}$)  & Node (ROI) \( i \) in the node (ROI) set $\mathcal{V} (\mathcal{R})$  \\
        \( \mathcal{E} \) & Edge set \\
        \( A \) & Adjacency matrix, \( A = [a_{ij}] \in \{0,1\}^{|\mathcal{V}| \times |\mathcal{V}|} \) \\
        \( \mathbf{X} \) & Node feature matrix, \( \mathbf{X} \in \mathbb{R}^{|\mathcal{V}| \times d} \) \\
        \( \mathbf{E} \) & Edge feature matrix, \( \mathbf{E}  \in \mathbb{R}^{|\mathcal{E}| \times d^\prime} \)\\ 
        \( d/d^\prime \) & Node/edge feature dimension \\
        \( \mathbf{T}_r (\mathbf{Z}_r) \) & Temporal (Normalized) BOLD signal for ROI \( r \), \( \mathbf{T}_r \in \mathbb{R}^T \) \\
        \( \mu_r (\sigma_r) \) & Mean (Standard deviation) of the BOLD signal for ROI \( r \) \\
        \( \Theta(t; \mathbf{Z}_r; \theta, \gamma) \) & Thresholding function for high-amplitude signal retention \\
        \( \theta (\gamma) \) & Threshold (Binarization) param. for high-amplitude 
        retention \\
        \hline
    \end{tabular}
    \vspace{-3ex}
\end{table}

\section{Correlation-based Edge Construction Details}\label{sec:correlation}
\vspace{0.5ex}
Here we formally define Pearson Correlation Coefficient, Spearman Correlation, and Kendall's Tau, in the context of fMRI neuroimaging data, i.e., signal relations between ROIs. 

\vspace{1ex}
\noindent \textit{Pearson Correlation Coefficient} between ROI $r_i$ and $r_j$ is defined as: 
\begin{equation}
  \rho_{r_i,r_j} = \frac{\text{cov}({\bf T}_{r_i},{\bf T}_{r_j})}{\sigma_{r_i} \sigma_{r_j}}
  \label{eq:pearson}
\end{equation}
where $\sigma_x$ and $\sigma_y$ correspond respectively to the standard deviation of signal ${\bf T}_{r_i}$ and ${\bf T}_{r_j}$, respectively, and defined in Equation~\ref{eq:stdev}. We note for simplicity here we use ${\bf T}_{r_i}$ to denote the temporal BOLD signal of ROI $r_i$, but is similarly defined if combining with the high-amplitude signal retention process defined in Section~\ref{sec:denoising}. 

\vspace{1ex}
\noindent \textit{Spearman Correlation} is a non-parametric rank correlation that measures the strength and direction of a monotonic relationship between two variables~\cite{ahmadi2023comparative}. It first ranks the data and is then calculated thereafter following the same method in Equation~\ref{eq:pearson}.

\vspace{1ex}
\noindent \textit{Kendall's Tau} is another widely used non-parametric method for computing the correlation between two variables~\cite{ahmadi2023comparative}. 
It evaluates 
the probability a randomly chosen pair of observations will have the same order in both variables. 
When applied to two signals, ${\bf T}_{r_i}$ and ${\bf T}_{r_j}$ (for regions \( r_i \) and \( r_j \)), 
Kendall’s Tau quantifies the degree of ordinal association between them. Specifically, for each pair of time points \( (t_a, t_b) \), where \( a < b \), the pair is considered concordant if the relative ordering of the signals \( T_{r_i}(t_a) \), \( T_{r_j}(t_a) \) is the same as that of \( T_{r_i}(t_b) \), \( T_{r_j}(t_b) \). Conversely, the pair is discordant if the order is reversed.
This 
ensures that it captures the consistency of rank ordering between the signals \( {\bf T}_{r_i} \) and \( {\bf T}_{r_j} \) over time, making it a robust and interpretable alternative to Pearson’s and Spearman’s correlation coefficients. Kendall's Tau is calculated as follows:
\begin{equation}
  \tau_{r_i,r_j} = \frac{n_c(r_i,r_j) - n_d(r_i,r_j)}{\frac{1}{2} n (n-1)}
  \label{eq:kendall_tau}
\end{equation}
where $n_c(r_i,r_j)$ and $n_d(r_i,r_j)$ correspond to the number of concordant and discordant pairs between $r_i$ and $r_j$, having $n$ pairs. 


\section{Additional Details on fMRI to Graph Construction}\label{sec:standard_pipeline}

The most common approach for constructing a brain graph from fMRI data is visualized in Figure~\ref{fig:pipeline_img} and involves the following steps: First, the fMRI data must be carefully preprocessed to remove known artifacts and noise. After preprocessing, the fMRI signals are extracted and averaged within the same ROIs, which are pre-defined based on a brain atlas template. Next, pairwise correlation coefficients are calculated for each pair of ROIs to construct a functional connectivity (FC) matrix based on their representive z-normalized time series. The Pearson correlation coefficient is the most commonly used measure in previous research. This FC matrix is often used as the node feature in the graph. Finally, edges are defined by applying a threshold to the FC matrix, representing strong FC in the brain graph while avoiding an undesired fully connected graph~\cite{said2023neurograph}. In our study, we take top 5\% positive correlation value in the functional connectivity matrix as the edges in the graph.

\section{Dataset Details}\label{sec:appendix_dataset}

\textbf{Human Connectome Project (HCP) S1200}~\cite{van2013wu}: This publicly available neuroimaging dataset was collected from healthy young adult participants and includes both resting-state and task-based fMRI data. For the resting-state dataset, participants remained at rest during the scan, and their demographics served as labels for graph classification. We used resting-state fMRI data from 1,078 subjects, excluding those with fewer than 1,200 frames. Our methods were applied to the task of gender classification, where each subject's gender served as the label (2 classes). For the task dataset, participants performed specific cognitive or motor tasks during the scan. The dataset includes over 7,000 scans, covering seven task types: working memory, social cognition, relational processing, motor function, language, gambling, and emotion. The type of task performed served as the label for graph classification (7 classes). For preprocessing, we used the Minimally Preprocessed HCP Pipeline for all scans~\cite{glasser2013minimal}, followed by additional steps to enhance data quality. We applied brain parcellation using the Schaefer atlas~\cite{schaefer2018local} to extract the mean fMRI timeseries for each region of interest (ROI). To reduce instrumental noise, we removed scanner drifts by regressing out linear and quadratic trends. Motion artifacts were mitigated using rigid-body motion regression, removing six motion parameters and their derivatives before graph construction.

\vspace{1ex}
\noindent \textbf{Autism Brain Imaging Data Exchange (ABIDE)} ~\cite{di2014autism}: A collaborative initiative involving 17 international research sites, aimed at enhancing our understanding of Autism Spectrum Disorder (ASD). ABIDE provides resting-state fMRI data from individuals with and without ASD. We used data from 1,009 subjects, with binary labels indicating ASD diagnosis. For preprocessing, we applied the C-PAC pipeline~\cite{craddock2012whole}, including brain parcellation, slice timing correction, motion correction, and spatial normalization to the MNI152 template before graph construction.

\section{Detailed Result Tables}\label{sec:appendix_table}
\suppressfloats[t]
In Tables~\ref{tab:denoising_abide_full} and \ref{tab:corr_abide_full} we report the more detailed results of Tables~\ref{tab:denoising_abide} and \ref{tab:corr_abide}, respectively. 
We also provide the full detailed results tables for selected lagged correlation experiments, specifically Tables~\ref{tab:8} and 
\ref{tab:10}. 
Finally, the detailed results for the edge featurization are provided in Tables~\ref{tab:edge_feature} and \ref{tab:edge_feature-abide}. 

\begin{table*}[h]
\centering
\caption{Classification performance of GNNs on ABIDE dataset under different feature selection strategies and ROI settings.}
\label{tab:denoising_abide_full}
\resizebox{\textwidth}{!}{
\begin{tabular}{lcccccccccc}
\toprule
& \multicolumn{4}{c}{\textbf{ROI 200}} & \multicolumn{4}{c}{\textbf{ROI 400}} & \textbf{Avg. across ROIs \& GNNs} \\
\cmidrule(lr){2-5} \cmidrule(lr){6-9}
& \textbf{GCN} & \textbf{GAT} & \textbf{SAGE} & \textbf{Avg. across GNNs} & \textbf{GCN} & \textbf{GAT} & \textbf{SAGE} & \textbf{Avg. across GNNs} & \\
\midrule
\textbf{Baseline} & 60.00 ± 2.94 & 62.13 ± 2.69 & 62.32 ± 1.67 & 61.48 & 62.32 ± 3.25 & 64.06 ± 1.61 & 63.86 ± 2.15 & 63.41 & 62.45 \\
\textbf{\(\Theta_{\theta_{p30}\gamma_0}\)} & 61.74 ± 3.14 & 62.13 ± 1.58 & 62.22 ± 2.19 & 62.03 & 61.64 ± 2.84 & 63.09 ± 2.49 & 57.87 ± 4.30 & 60.87 & 61.45 \\
\textbf{\(\Theta_{\theta_{p30}\gamma_1}\)} & 58.94 ± 6.11 & 61.45 ± 4.19 & 61.35 ± 2.63 & 60.58 & 62.13 ± 1.82 & 63.96 ± 1.13 & 60.00 ± 5.29 & 62.03 & 61.31 \\
\textbf{\(\Theta_{\theta_{sd1}\gamma_0}\)} & 62.02 ± 2.22 & 62.32 ± 6.70 & 61.72 ± 3.28 & 62.02 & 60.77 ± 4.24 & 62.73 ± 1.17 & 61.53 ± 4.05 & 61.68 & 61.85 \\
\textbf{\(\Theta_{\theta_{sd1}\gamma_1}\)} & 65.86 ± 2.11 & 63.03 ± 4.86 & 63.84 ± 2.36 & 64.24 & 61.64 ± 1.27 & 61.75 ± 0.91 & 62.08 ± 2.71 & 61.82 & 63.03 \\
\bottomrule
\end{tabular}
}
\end{table*}

\begin{table*}[h]
\centering
\caption{Classification performance of GNNs on ABIDE dataset under different correlation strategies and ROI settings.}
\label{tab:corr_abide_full}
\resizebox{\textwidth}{!}{
\begin{tabular}{lcccccccccc}
\toprule
& \multicolumn{4}{c}{\textbf{ROI 200}} & \multicolumn{4}{c}{\textbf{ROI 400}} & \textbf{Avg. across ROIs \& GNNs} \\
\cmidrule(lr){2-5} \cmidrule(lr){6-9}
& \textbf{GCN} & \textbf{GAT} & \textbf{SAGE} & \textbf{Avg. across GNNs} & \textbf{GCN} & \textbf{GAT} & \textbf{SAGE} & \textbf{Avg. across GNNs} & \\
\midrule
\textbf{Baseline} & 60.00 ± 2.94 & 62.13 ± 2.69 & 62.32 ± 1.67 & 61.48 & 62.32 ± 3.25 & 64.06 ± 1.61 & 63.86 ± 2.15 & 63.41 & 62.45 \\
\textbf{Spearman} & 61.93 ± 1.77 & 65.22 ± 2.66 & 60.87 ± 2.03 & 62.67 & 63.00 ± 3.37 & 63.29 ± 1.53 & 63.48 ± 2.83 & 63.26 & 62.97 \\
\textbf{Kendall} & 62.13 ± 3.03 & 63.09 ± 1.48 & 61.06 ± 1.87 & 62.09 & 63.77 ± 1.76 & 63.77 ± 1.65 & 63.38 ± 2.13 & 63.64 & 62.87 \\
\bottomrule
\end{tabular}
}
\end{table*}

\begin{table*}[h]
\centering
\caption{Classification performance of GNNs on HCP-Rest dataset under different lagged correlation values and ROI settings.}
\label{tab:8}
\resizebox{\textwidth}{!}{
\begin{tabular}{lccccccccccccc}
\toprule
& \multicolumn{4}{c}{\textbf{ROI 100}} & \multicolumn{4}{c}{\textbf{ROI 400}} & \multicolumn{4}{c}{\textbf{ROI 1000}} & \textbf{Avg. across ROIs \& GNNs} \\
\cmidrule(lr){2-5} \cmidrule(lr){6-9} \cmidrule(lr){10-13}
& \textbf{GCN} & \textbf{GAT} & \textbf{SAGE} & \textbf{Avg. across GNNs} & \textbf{GCN} & \textbf{GAT} & \textbf{SAGE} & \textbf{Avg. across GNNs} & \textbf{GCN} & \textbf{GAT} & \textbf{SAGE} & \textbf{Avg. across GNNs} & \\
\midrule
\textbf{Baseline} & 74.07 ± 1.13 & 74.35 ± 2.56 & 71.11 ± 3.76 & 73.18 & 74.63 ± 0.68 & 76.02 ± 0.80 & 75.37 ± 2.08 & 75.34 & 81.39 ± 2.81 & 78.33 ± 3.43 & 81.48 ± 1.80 & 80.4 & 76.31 \\
\textbf{Lag = 1} & 72.13 ± 3.43 & 71.30 ± 2.21 & 72.69 ± 1.17 & 72.04 & 75.56 ± 2.22 & 74.91 ± 3.67 & 75.19 ± 1.12 & 75.22 & 78.33 ± 8.95 & 80.09 ± 2.41 & OOM & 79.21 & 75.03 \\
\textbf{Lag = 2} & 76.39 ± 2.57 & 73.06 ± 2.86 & 73.15 ± 2.01 & 74.2 & 76.94 ± 1.26 & 76.48 ± 2.58 & 76.30 ± 0.99 & 76.57 & 78.15 ± 10.90 & 80.56 ± 1.94 & OOM & 79.36 & 76.38 \\
\textbf{Lag = 3} & 74.17 ± 3.42 & 73.89 ± 1.96 & 74.07 ± 1.80 & 74.04 & 77.96 ± 1.57 & 75.56 ± 1.26 & 78.43 ± 2.10 & 77.32 & 84.91 ± 1.65 & 82.69 ± 0.86 & OOM & 83.8 & 77.71 \\
\textbf{Lag = 4} & 73.06 ± 1.81 & 73.06 ± 2.49 & 73.52 ± 1.42 & 73.21 & 79.81 ± 2.32 & 78.89 ± 1.54 & 78.70 ± 0.97 & 79.13 & 84.72 ± 0.65 & 83.24 ± 3.23 & OOM & 83.98 & 78.13 \\
\textbf{Lag = 5} & 75.37 ± 1.29 & 73.33 ± 1.33 & 74.63 ± 1.79 & 74.44 & 81.85 ± 1.74 & 79.26 ± 0.90 & 80.46 ± 1.35 & 80.52 & 84.07 ± 1.48 & 85.28 ± 2.04 & OOM & 84.68 & 79.28 \\
\textbf{Lag = 6} & 73.80 ± 3.77 & 74.91 ± 1.42 & 74.44 ± 1.72 & 74.38 & 80.00 ± 2.24 & 80.46 ± 2.51 & 79.07 ± 0.80 & 79.84 & 83.24 ± 3.25 & 82.78 ± 2.06 & OOM & 83.01 & 78.59 \\
\textbf{Lag = 7} & 73.52 ± 1.79 & 75.46 ± 1.52 & 73.33 ± 2.51 & 74.1 & 78.80 ± 2.18 & 81.11 ± 1.64 & 79.07 ± 1.50 & 79.66 & 85.09 ± 1.45 & 84.26 ± 1.13 & OOM & 84.68 & 78.83 \\
\textbf{Lag = 8} & 73.80 ± 1.62 & 75.46 ± 1.99 & 73.89 ± 0.81 & 74.38 & 79.91 ± 1.33 & 78.24 ± 1.28 & 77.96 ± 1.45 & 78.7 & 84.26 ± 1.37 & 84.44 ± 1.51 & OOM & 84.35 & 78.5 \\
\textbf{Lag = 9} & 74.26 ± 2.39 & 72.69 ± 1.68 & 73.80 ± 2.04 & 73.58 & 77.41 ± 1.84 & 77.78 ± 1.40 & 76.02 ± 0.61 & 77.07 & 81.85 ± 2.24 & 84.54 ± 1.36 & OOM & 83.2 & 77.29 \\
\textbf{Lag = 10} & 72.96 ± 1.67 & 75.19 ± 1.93 & 71.85 ± 2.24 & 73.33 & 75.83 ± 0.54 & 77.41 ± 1.07 & 76.30 ± 3.59 & 76.51 & 82.41 ± 1.40 & 83.43 ± 1.07 & OOM & 82.92 & 76.92 \\
\bottomrule
\end{tabular}
}
\end{table*}

\begin{table*}[h]
\centering
\caption{Classification performance of GNNs on ABIDE dataset under different lagged correlation values and ROI settings.}
\label{tab:10}
\resizebox{\textwidth}{!}{%
\begin{tabular}{lccccccccc}
\toprule
& \multicolumn{4}{c}{\textbf{ROI 200}} & \multicolumn{4}{c}{\textbf{ROI 400}} & \textbf{Avg. across ROIs \& GNNs} \\
\cmidrule(lr){2-5} \cmidrule(lr){6-9}
& \textbf{GCN} & \textbf{GAT} & \textbf{SAGE} & \textbf{Avg. across GNNs} & \textbf{GCN} & \textbf{GAT} & \textbf{SAGE} & \textbf{Avg. across GNNs} & \\
\midrule
\textbf{Baseline} & 60.00 ± 2.94 & 62.13 ± 2.69 & 62.32 ± 1.67 & 61.48 & 62.32 ± 3.25 & 64.06 ± 1.61 & 63.86 ± 2.15 & 63.41 & 62.45 \\
\textbf{Lag = 1} & 63.96 ± 1.58 & 64.64 ± 1.20 & 62.13 ± 2.04 & 63.58 & 65.51 ± 2.97 & 61.16 ± 2.74 & 65.89 ± 4.09 & 64.19 & 63.88 \\
\textbf{Lag = 2} & 61.35 ± 1.88 & 57.39 ± 2.49 & 58.45 ± 2.57 & 59.06 & 61.55 ± 3.17 & 60.77 ± 5.12 & 66.67 ± 5.37 & 63.00 & 61.03 \\
\textbf{Lag = 3} & 59.52 ± 3.58 & 57.29 ± 3.66 & 58.94 ± 4.12 & 58.58 & 61.74 ± 4.43 & 63.19 ± 7.18 & 65.12 ± 4.69 & 63.35 & 60.97 \\
\textbf{Lag = 4} & 58.26 ± 5.01 & 57.78 ± 2.78 & 60.68 ± 2.71 & 58.91 & 63.96 ± 3.45 & 63.29 ± 4.39 & 64.15 ± 4.44 & 63.80 & 61.35 \\
\textbf{Lag = 5} & 60.68 ± 2.47 & 61.84 ± 2.57 & 62.13 ± 1.95 & 61.55 & 64.64 ± 2.67 & 67.15 ± 4.81 & 62.51 ± 3.85 & 64.77 & 63.16 \\
\bottomrule
\end{tabular}
}
\end{table*}

\begin{table*}[h]
\centering
\caption{Classification results on HCP-Rest and HCP-Task with and without edge features.}
\label{tab:edge_feature}
\resizebox{\textwidth}{!}{%
\begin{tabular}{clccccccccccccc}
\toprule
\multirow{2}{*}{\textbf{Dataset}} &  & \multicolumn{4}{c}{\textbf{ROI 100}} & \multicolumn{4}{c}{\textbf{ROI 400}} & \multicolumn{4}{c}{\textbf{ROI 1000}} & \textbf{Avg. across}  \\
\cmidrule(lr){3-6} \cmidrule(lr){7-10} \cmidrule(lr){11-14}
&  & \textbf{GCN} & \textbf{GAT} & \textbf{SAGE} & \textbf{Avg. GNNs} & \textbf{GCN} & \textbf{GAT} & \textbf{SAGE} & \textbf{Avg. GNNs} & \textbf{GCN} & \textbf{GAT} & \textbf{SAGE} & \textbf{Avg. GNNs} & \textbf{ROIs \& GNNs} \\
\midrule
\multirow{2}{*}{\textbf{HCP-Rest}} 
& \textbf{Baseline} & 74.07 ± 1.13 & 74.35 ± 2.56 & 71.11 ± 3.76 & 73.18 & 74.63 ± 0.68 & 76.02 ± 0.80 & 75.37 ± 2.08 & 75.34 & 81.39 ± 2.81 & 78.33 ± 3.43 & 81.48 ± 1.80 & 80.40 & 76.31 \\
& \textbf{w/ Edge Feat.} & 70.22 ± 1.33 & 69.29 ± 1.43 & 71.30 ± 0.38 & 70.27 & 75.46 ± 2.27 & 77.62 ± 2.28 & 75.00 ± 2.73 & 76.03 & 80.56 ± 1.31 & 83.64 ± 3.86 & 81.64 ± 1.22 & 81.95 & 76.08 \\
\midrule
\multirow{2}{*}{\textbf{HCP-Task}} 
& \textbf{Baseline} & 91.05 ± 0.29 & 91.93 ± 0.50 & 91.71 ± 0.30 & 91.56 & 94.57 ± 0.60 & 95.08 ± 0.51 & 95.51 ± 0.19 & 95.05 & 96.06 ± 0.36 & 96.26 ± 0.62 & 96.49 ± 0.28 & 96.27 & 94.30 \\
& \textbf{w/ Edge Feat.} & 91.22 ± 0.98 & 92.05 ± 1.63 & 92.93 ± 0.43 & 92.07 & 94.09 ± 0.58 & 95.46 ± 0.28 & 95.90 ± 0.33 & 95.15 & OOM & OOM & OOM & / & 93.61 \\
\bottomrule
\end{tabular}%
}
\end{table*}

\begin{table*}[h]
\centering
\caption{Classification results on ABIDE with and without edge features.}
\label{tab:edge_feature-abide}
\resizebox{\textwidth}{!}{%
\begin{tabular}{lccccccccc}
\toprule
\multirow{2}{*}{} & \multicolumn{4}{c}{\textbf{ROI 200}} & \multicolumn{4}{c}{\textbf{ROI 400}} & \textbf{Avg. across ROIs \& GNNs} \\
\cmidrule(lr){2-5} \cmidrule(lr){6-9}
& \textbf{GCN} & \textbf{GAT} & \textbf{SAGE} & \textbf{Avg. across GNNs} & \textbf{GCN} & \textbf{GAT} & \textbf{SAGE} & \textbf{Avg. across GNNs} & \\
\midrule
\textbf{Baseline} & 60.00 ± 2.94 & 62.13 ± 2.69 & 62.32 ± 1.67 & 61.48 & 62.32 ± 3.25 & 64.06 ± 1.61 & 63.86 ± 2.15 & 63.41 & 62.45 \\
\textbf{w/ Edge Feat} & 61.84 ± 0.79 & 61.03 ± 2.28 & 60.39 ± 2.20 & 61.09 & 63.29 ± 3.88 & 64.09 ± 0.46 & 61.67 ± 2.41 & 63.02 & 62.05 \\
\bottomrule
\end{tabular}%
}
\end{table*}

\end{document}